\documentclass[runningheads]{llncs}

\makeatletter
\@namedef{ver@everyshi.sty}{}
\makeatother

\usepackage[utf8]{inputenc}

\usepackage{graphicx}
\usepackage{comment}
\usepackage{amsmath,amssymb} 
\usepackage{color}

\usepackage{eso-pic}
\usepackage[width=122mm,left=12mm,paperwidth=146mm,height=193mm,top=12mm,paperheight=217mm]{geometry}

\usepackage{lscape}
\usepackage{placeins}

\usepackage{bbm, dsfont}
\usepackage{relsize}

\usepackage{xspace}
\usepackage{caption}
\usepackage{subcaption}
\usepackage{float}

\usepackage{array}
\usepackage{ifthen}
\usepackage{booktabs}

\usepackage[dvipsnames]{xcolor}

\usepackage{tikz}
\usepackage{pgfplots}

\usepackage{microtype}

\makeatletter
\DeclareRobustCommand\onedot{\futurelet\@let@token\@onedot}
\def\@onedot{\ifx\@let@token.\else.\null\fi\xspace}
\def\eg{\emph{e.g}\onedot} 
\def\ie{\emph{i.e}\onedot} 
\def\cf{\emph{c.f}\onedot}

\def\etal{\emph{et\ al}\onedot}

\DeclareMathOperator*{\argmax}{argmax}

\DeclareMathOperator*{\softargmax}{softargmax}

\graphicspath{{./figures/}}

\definecolor{cb1}{RGB}{0,0,0}
\definecolor{cb2}{RGB}{0,73,73}
\definecolor{cb3}{RGB}{0,146,146}
\definecolor{cb4}{RGB}{255,109,182}
\definecolor{cb5}{RGB}{255,182,119}
\definecolor{cb6}{RGB}{73,0,146}
\definecolor{cb7}{RGB}{0,109,219}
\definecolor{cb8}{RGB}{182,109,255}
\definecolor{cb9}{RGB}{109,182,255}
\definecolor{cb10}{RGB}{182,219,255}
\definecolor{cb11}{RGB}{146,0,0}
\definecolor{cb12}{RGB}{146,73,0}
\definecolor{cb13}{RGB}{219,209,0}
\definecolor{cb14}{RGB}{36,255,36}
\definecolor{cb15}{RGB}{255,255,109}

\newcommand\plotlabel[3]{
\ifthenelse{\equal{#2}{x}}
    {\def\msz{5pt}}
    {\def\msz{3pt}}%
\begin{tikzpicture}[baseline]
    \begin{axis}[hide axis, scale only axis, height=2ex, width=9.5ex]
        \addplot [color=#1,line width=0.5mm,#3] coordinates {(0,0) (2,0)};
        \addplot [color=#1,line width=0.5mm,mark size=\msz,mark=#2] coordinates {(1,0)};
    \end{axis}
\end{tikzpicture}
}

\newcommand\plotlabelv[3]{
\ifthenelse{\equal{#2}{x}}
    {\def\msz{5pt}}
    {\def\msz{3pt}}%
\begin{tikzpicture}[baseline]
    \begin{axis}[hide axis, scale only axis, height=2ex, width=7ex]
        \addplot [color=#1,line width=0.5mm,#3] coordinates {(0,0) (2,0)};
        \addplot [color=#1,line width=0.5mm,mark size=\msz,mark=#2] coordinates {(1,0)};
    \end{axis}
\end{tikzpicture}\hspace{-0.5em}
}

\usepackage{etoolbox}
\makeatletter
\patchcmd{\maketitle}
 {\def\@makefnmark}
 {\def\@makefnmark{}\def\useless@macro}
 {}{}
\makeatother

\begin{document}
\pagestyle{headings}
\mainmatter
\def\ECCVSubNumber{X}

\title{Efficient Neighbourhood Consensus Networks via Submanifold Sparse Convolutions}

\titlerunning{Sparse-NCNet: Efficient Neighbourhood Consensus Networks}

\author{Ignacio Rocco$^{1,2}$ \qquad Relja Arandjelovi\'{c}$^{3}$ \qquad Josef Sivic$^{1,2,4}$}
\institute{$^{1}$Inria \qquad $^{2}$DI-ENS\thanks{WILLOW project, Département d’informatique, École Normale Supérieure, CNRS, PSL Research University, Paris, France.} \qquad  $^{3}$DeepMind \qquad $^{4}$CIIRC\thanks{Czech Institute of Informatics, Robotics and Cybernetics at the Czech Technical University in Prague.} \\ \url{http://www.di.ens.fr/willow/research/sparse-ncnet/}}

\authorrunning{Ignacio Rocco, Relja Arandjelovi\'{c} and Josef Sivic}

\maketitle

\begin{abstract}
In this work we target the problem of estimating accurately localised correspondences between a pair of images. We adopt the recent Neighbourhood Consensus Networks that have demonstrated promising performance for difficult correspondence problems and propose modifications to overcome their main limitations: large memory consumption, large inference time and poorly localised correspondences. Our proposed modifications can reduce the memory footprint and execution time more than $10\times$, with equivalent results. This is achieved by \emph{sparsifying} the correlation tensor containing tentative matches, and its subsequent processing with a 4D CNN using submanifold sparse convolutions. Localisation accuracy is significantly improved by processing the input images in higher resolution, which is possible due to the reduced memory footprint, and by a novel two-stage correspondence relocalisation module. The proposed Sparse-NCNet method obtains state-of-the-art results on the HPatches Sequences and InLoc visual localisation benchmarks, and competitive results in the Aachen Day-Night benchmark.

\end{abstract}

\section{Introduction} 
Finding correspondences between images depicting the same 3D scene is one of the fundamental tasks in computer vision~\cite{julesz1962towards,marr1976cooperative,mori1973iterative} with applications in 3D reconstruction~\cite{schoenberger2016sfm,schoenberger2016mvs,widya2018structure}, visual localisation~\cite{Germain19,sattler2018benchmarking,Taira18} or pose estimation~\cite{gao2003complete,Grabner18,persson2018lambda}. The predominant approach currently consists of first \emph{detecting} salient local features, by selecting the local extrema of some form of
feature selection function, and then \emph{describing} them by some form of feature descriptor~\cite{SURF,lowe2004distinctive,ORB}. While hand-crafted features such as Hessian affine detectors~\cite{mikolajczyk2002affine} with SIFT descriptors~\cite{lowe2004distinctive} have obtained impressive performance under strong viewpoint changes and constant illumination~\cite{mikolajczyk2005comparison}, their robustness to illumination changes is limited~\cite{mikolajczyk2005comparison,feats_day_night}. More recently, a variety of trainable keypoint detectors~\cite{laguna2019keynet,Lenc16,Mishkin2018AffNet,verdie2015tilde} and descriptors~\cite{pn_net,TFeat,han2015matchnet,HardNet,L2Net,deepcompare} have been proposed, with the purpose of obtaining increased robustness over hand-crafted methods. While this approach has achieved some success, extreme illumination changes such as day-to-night matching combined with changes in camera viewpoint remain a challenging open problem~\cite{visuallocalizationbenchmark,Dusmanu2019CVPR,Germain19}. In particular, all local feature methods, whether hand-crafted or trained, suffer from missing detections under these extreme appearance changes. 

\begin{figure}[t]
    \centering
    \begin{subfigure}[b]{0.33\textwidth}
        \includegraphics[width=\textwidth]{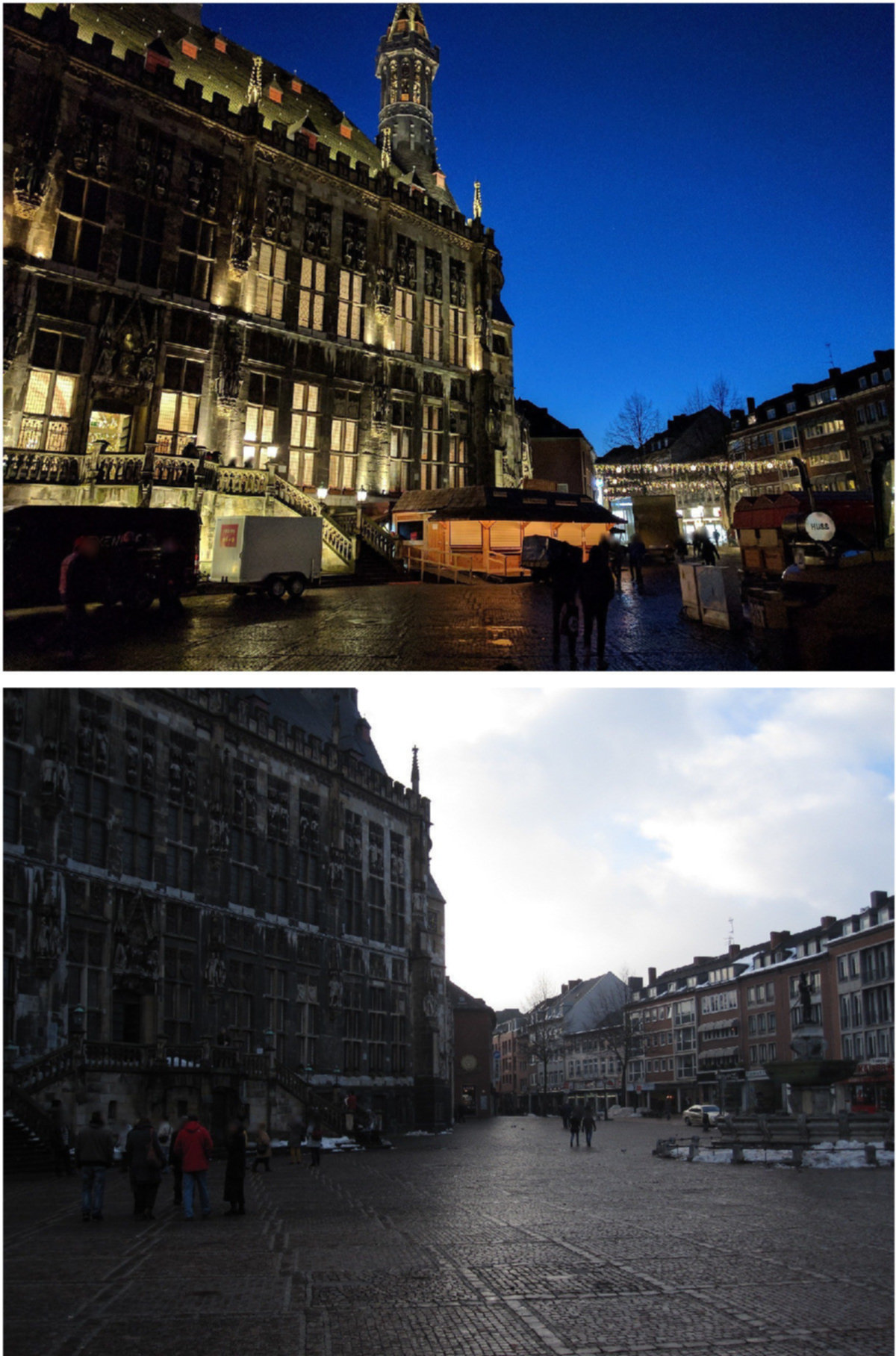}
        \caption{Input images}
    \end{subfigure}
    \hspace{-0.5em}
    \begin{subfigure}[b]{0.33\textwidth}
        \includegraphics[width=\textwidth]{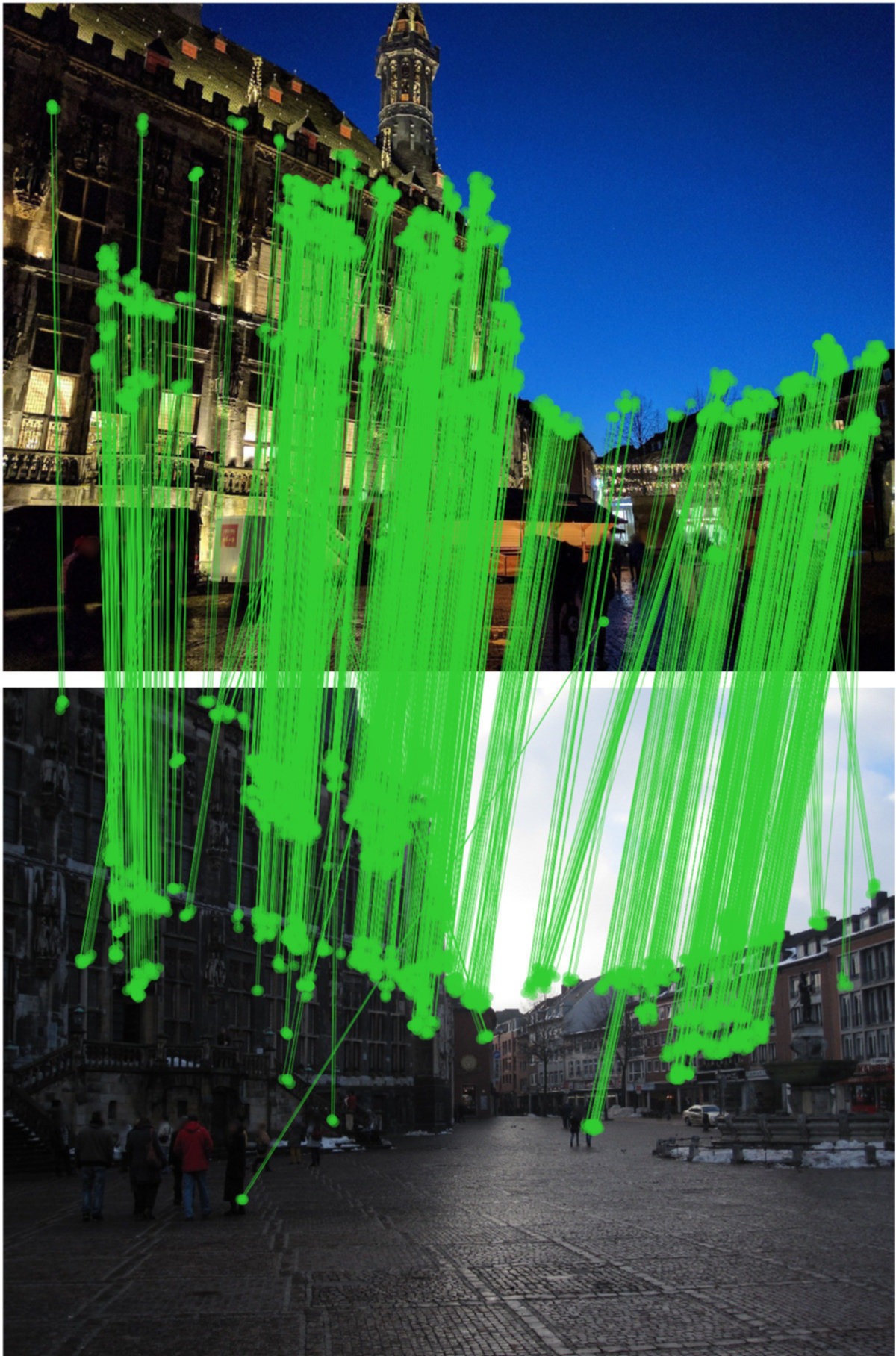}
        \caption{Output matches}
    \end{subfigure}
    \hspace{-0.5em}
    \begin{subfigure}[b]{0.33\textwidth}
        \includegraphics[width=\textwidth]{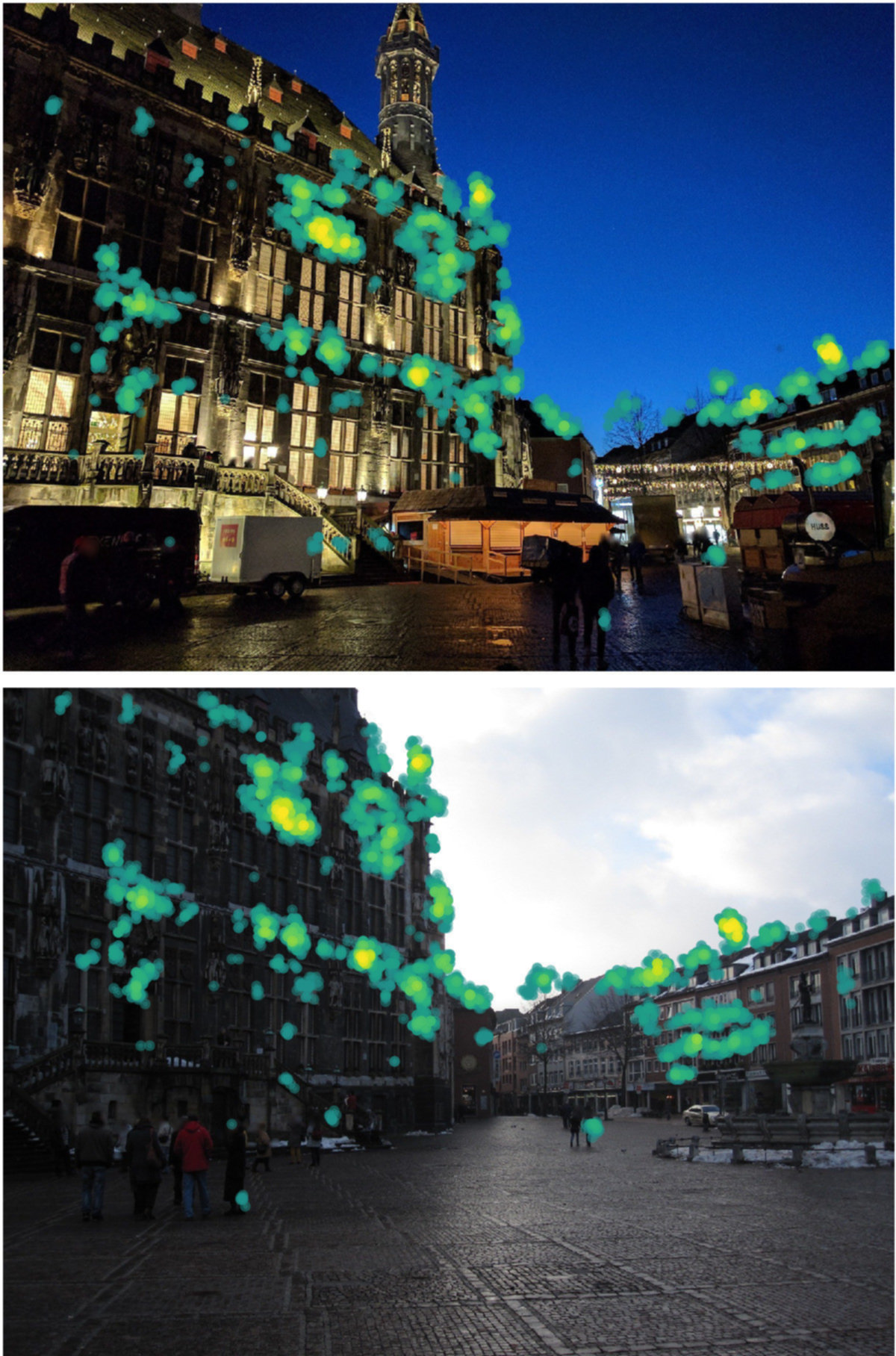}
        \caption{Match confidence}
    \end{subfigure}
    \caption{\textbf{Correspondence estimation with Sparse-NCNet.} Given an input image pair (a), we show the \emph{raw} output correspondences produced by Sparse-NCNet (b) which contain groups of spatially coherent matches. These groups tend to form around highly-confident matches, which are shown in yellow shades (c) (see Appendix~\ref{sec:intuitions} for a discussion on this behaviour and additional examples).
    }
    \label{fig:teaser}
\end{figure}

In order to overcome this issue, the detection stage can be avoided and, instead, features can be extracted on a dense grid across the image. This approach has been successfully used for both place recognition~\cite{Arandjelovic2016,Germain19,noh2017DELF,Torii2015_24_7} and image matching~\cite{NCNet,sattler2018benchmarking,widya2018structure}. However, extracting features densely comes with additional challenges: it is memory intensive and the localisation accuracy of the features is limited by the sampling interval of the grid used for the extraction.

In this work we adopt the dense feature extraction approach. In particular, we build on the recent Neighbourhood Consensus Networks (NCNet)~\cite{NCNet}, 
that allow for jointly trainable feature extraction, matching, and match-filtering to directly output a strong set of (mostly) correct correspondences. Our proposed approach, Sparse-NCNet, seeks to overcome the limitations of the original NCNet formulation, namely: large memory consumption, high execution time and poorly localised correspondences. 

Our contributions are the following. First, we propose the efficient Sparse-NCNet model, which is based on a 4D convolutional neural network operating on a \emph{sparse} correlation tensor, which is obtained by storing only the most promising correspondences, instead of
the set of all possible correspondences. Sparse-NCNet processes this sparse correlation tensor with submanifold sparse convolutions~\cite{3DSemanticSegmentationWithSubmanifoldSparseConvNet} and can obtain equivalent results to NCNet while being several times faster (up to $10\times$) and requiring much less memory (up to $20\times$) without decrease in performance compared to the original NCNet model. Second, we propose a two-stage relocalisation module to improve the localisation accuracy of the correspondences output by Sparse-NCNet. Finally, we show that the proposed model significantly outperforms state-of-the-art results on the HPatches Sequences~\cite{hpatches2017cvpr} benchmark for image matching with challenging viewpoint and illumination changes and the InLoc~\cite{Taira18} benchmark for indoor localisation and camera pose estimation. Furthermore, we show our model obtains competitive results on the Aachen Day-Night benchmark~\cite{sattler2018benchmarking}, which evaluates day-night feature matching for the task of camera localisation. 
An example of the correspondences produced by our method is presented in Fig.~\ref{fig:teaser}. \textbf{Our code and models are available online\footnote{\url{\url{http://www.di.ens.fr/willow/research/sparse-ncnet/}}}.}

\section{Related work}
In this section, we review the relevant related work. 

\paragraph{Matching with trainable local features.}
Most recent work in trainable local features has focused on learning more robust keypoint \emph{descriptors}~\cite{pn_net,TFeat,han2015matchnet,HardNet,L2Net,deepcompare}. Initially these descriptors were used in conjunction with classic hand-crafted keypoint detectors, such as DoG~\cite{lowe2004distinctive}. Recently, trainable keypoint \emph{detectors} where also proposed~\cite{laguna2019keynet,Lenc16,Mishkin2018AffNet,verdie2015tilde}, as well as methods providing both \emph{detection and description}~\cite{Detone2018CVPRW,Dusmanu2019CVPR,ono2018lf,r2d2,yi2016lift}. From these, some adopt the classic approach of first performing detection on the whole image and then computing descriptors from local image patches, cropped around the detected keypoints~\cite{ono2018lf,yi2016lift}, while the most recent methods compute a joint representation from which both detections and descriptors are computed~\cite{Detone2018CVPRW,Dusmanu2019CVPR,r2d2}.
In most cases, local features obtained by these methods are independently matched using nearest-neighbour search with the Euclidean distance~\cite{pn_net,TFeat,HardNet,L2Net}, although some works have proposed to learn the distance function as well~\cite{han2015matchnet,deepcompare}.
As discussed in the previous section, local features are prone to loss of detections under extreme lighting changes~\cite{Germain19}. In order to alleviate this issue, in this work we adopt the usage of dense features, which are described next.

\paragraph{Matching with densely extracted features.}
Motivated by applications in large-scale visual search, others have found that using densely extracted features provides additional robustness to illumination changes compared to local features extracted at detected keypoints, which suffer from low repeatability under strong illumination changes~\cite{Torii2015_24_7,zhao2013oriented}.
This approach was also adopted by later work~\cite{Arandjelovic2016,noh2017DELF}. Such densely extracted features used for image retrieval are typically computed on a coarse low resolution grid (\eg $40\times30)$. However, such coarse localisation of the dense features is not an issue for visual retrieval, as the dense features are not directly matched, but rather aggregated into a single image-level descriptor, which is used for retrieval.
Recently, densely extracted features have been also employed directly for 3D computer vision tasks, such as 3D reconstruction~\cite{widya2018structure}, indoor localisation and camera pose estimation~\cite{Taira18}, and outdoor localisation with night queries~\cite{Germain19,sattler2018benchmarking}. In these methods, correspondences are obtained by nearest-neighbour search performed on extracted descriptors, and filtered by the mutual nearest-neighbour criterion~\cite{oron2017best}.
In this work, we build on the NCNet method~\cite{NCNet}, where the match filtering function is learnt from data. Different recent methods for learning to filter matches are discussed next.

\paragraph{Learning to filter incorrect matches.}
When using both local features extracted at keypoints or densely extracted features, the obtained matches by nearest-neighbour search contain a certain portion of incorrect matches.
In the case of local features, a heuristic approach such as Lowe's ratio test~\cite{lowe2004distinctive} can be used to filter these matches. However the ratio threshold value needs to be manually tuned for each method. To avoid this issue, filtering by mutual nearest neighbours can be used instead~\cite{Dusmanu2019CVPR}.
Recently, trainable approaches have also been proposed for the task of filtering local feature correspondences~\cite{NG_RANSAC,PointCN,sarlin2019superglue,OANet}. Yi~\etal~\cite{PointCN} propose a neural-network architecture that operates on 4D match coordinates and classifies each correspondence as either correct or incorrect. Brachmann~\etal~\cite{NG_RANSAC} propose the Neural-guided RANSAC, which extends the previous method to produce weights instead of classification labels, which are used to guide RANSAC sampling. Zhang~\etal~\cite{OANet} also extend the work of Yi~\etal in their proposed Order-Aware Networks, which capture local context by clustering 4D correspondences onto a set of ordered clusters, and global context by processing these clusters with a multi-layer perceptron. Finally, Sarlin~\etal~\cite{sarlin2019superglue} describe a graph neural network followed by an optimisation procedure to estimate correspondences between two set of local features. These methods were specifically designed for filtering local features extracted at keypoint locations and not features extracted on a dense grid. Furthermore, these methods are focused only on learning match filtering, and are decoupled from the problem of learning how to detect and describe the local features.

In this paper we build on the NCNet method~\cite{NCNet} for filtering incorrect matches, which was designed for dense features. Furthermore, contrary to the above described methods, our approach  performs feature extraction, matching and match filtering in a single pipeline.

\paragraph{Improved feature localisation.}
Recent methods for local feature detection and description which use a joint representation~\cite{Detone2018CVPRW,Dusmanu2019CVPR} as well as methods for dense feature extraction~\cite{NCNet,widya2018structure} suffer from poor feature localisation, as the features are extracted on a low-resolution grid. Different approaches have been proposed to deal with this issue.
The D2-Net method~\cite{Dusmanu2019CVPR} follows the approach used in SIFT~\cite{lowe2004distinctive} for refining the keypoint positions, which consists of locally fitting a quadratic function to the feature detection function around the feature position and solving for the extrema.
The Superpoint method~\cite{Detone2018CVPRW} uses a CNN decoder that produces a one-hot output for each $8\times 8$ pixel cell of the input image (in case a keypoint is effectively detected in this region), therefore achieving pixel-level accuracy.
Others~\cite{widya2018structure} use the intermediate higher resolution features from the CNN to improve the feature localisation, by assigning to each pooled feature the position of the feature with highest L2 norm from the preceding higher resolution map (and which participated in the pooling). This process can be repeated up to the input image resolution.

The relocalisation approach of NCNet~\cite{NCNet} is based on a max-argmax operation on the 4D correlation tensor of exhaustive feature matches. This approach can only increase the resolution of the output matches by a factor of $2$. 
In contrast, we describe a new two-stage relocalisation module that builds on the approach used in NCNet, by combining a hard relocalisation stage that has similar effects to NCNet's max-argmax operation, with a soft-relocalisation stage that obtains sub-feature-grid accuracy via interpolation.

\paragraph{Sparse Convolutional Neural Networks} were recently introduced~\cite{graham2015sparse,graham2014spatially} for the purpose of processing sparse 2D data, such as handwritten characters~\cite{graham2014spatially}; 3D data, such as 3D point-clouds~\cite{graham2015sparse}; or even 4D data, such as temporal sequences of 3D point clouds~\cite{choy20194d}. These models have shown great success in 3D point-cloud processing tasks such as semantic segmentation~\cite{choy20194d,3DSemanticSegmentationWithSubmanifoldSparseConvNet} and point-cloud registration~\cite{FCGF2019,gojcic2020learning}. In this work, we use networks with \emph{submanifold sparse convolutions}~\cite{3DSemanticSegmentationWithSubmanifoldSparseConvNet} for the task of filtering correspondences between images, which can be represented as a sparse set of points in a 4D space of image coordinates. In submanifold sparse convolutions, the active sites remain constant between the input and output of each convolutional layer. As a result, the sparsity level remains fixed and does not change after each convolution operation. To the best of our knowledge this is the first time these models are applied to the task of match filtering.

\section{Sparse Neighbourhood Consensus Networks}
In this section we detail the proposed Sparse Neighbourhood Consensus Networks. We start with a brief review of Neighbourhood Consensus Networks~\cite{NCNet} identifying their main limitations. Next, we describe our approach which overcomes these limitations. 

\subsection{Review: Neighbourhood Consensus Networks}
The Neighbourhood Consensus Network~\cite{NCNet} is a method for feature extraction, matching and match filtering. Contrary to most methods, which operate on local features, NCNet operates on dense feature maps $(f^A, f^B)\in \mathbb{R}^{h\times w\times c}$ with $c$ channels, which are extracted over a regular grid of $h\times w$ spatial resolution. These are obtained from the input image pair $(I_A,I_B)\in \mathbb{R}^{H\times W\times 3}$
by a fully convolutional feature extraction network. The resolution $h\times w$ of the extracted dense features is typically $1/8$ or $1/16$ of the input image resolution $H\times W$, depending on the particular feature extraction network architecture used. 

Next, the exhaustive set of all possible matches between the dense feature maps $f^A$ and $f^B$ is computed and stored in a 4D correlation tensor $c^{AB}\in \mathbb{R}^{h\times w\times h \times w}$. Finally, the correspondences in $c^{AB}$ are filtered by a 4D CNN. This network can detect coherent spatial matching patterns and propagate information from the most certain matches to their neighbours, robustly identifying the correct correspondences. This last filtering step is inspired by the neighbourhood consensus procedure~\cite{bian2017gms,Schaffalitzky02a,schmid1997local,Sivic03,zhang1995robust}, where a particular match is verified by analysing the existence of other coherent matches in its spatial neighbourhood in both images.

Despite its promising results, the original formulation of Neighbourhood Consensus Networks has three main drawbacks that limit its practical application: it is (i) memory intensive, (ii) slow, and (iii) matches are poorly localised. These points are discussed in detail next.

\paragraph{High memory requirements.} The high memory requirements are due to the computation of the correlation tensor $c^{AB}\in \mathbb{R}^{h\times w\times h \times w}$ which stores all matches between the densely extracted image features $(f^A, f^B)\in \mathbb{R}^{h\times w\times c}$. Note that the number of elements in the correlation tensor ($h\times w\times h \times w$) grows quadratically with respect to the number of features ($h\times w$) of the dense feature maps $(f^A,f^B)$, therefore limiting the ability to increase the feature resolution. For instance, for dense feature maps of resolution $200\times 150$, the correlation tensor would require by itself $3.4$GB of GPU memory in the standard 32-bit float precision. Furthermore, processing this correlation tensor using the subsequent 4D CNN would require more than $50$GB of GPU memory, which is much more than what is currently available on most standard GPUs. While 16-bit half-float precision could be used to halve these memory requirements, they would still be prohibitively large. 

\paragraph{Long processing time.} In addition, Neighbourhood Consensus Networks are slow as the full dense correlation tensor must be processed.  For instance, processing the $100\times 75\times 100\times 75$ correlation tensor containing matches between a pair of dense feature maps of $100\times 75$ resolution takes approximately $10$ seconds on a standard Tesla T4 GPU.

\paragraph{Poor match localisation.} Finally, the high-memory requirements limit the maximum feature map resolution that can be processed, which in turn limits the localisation accuracy of the estimated correspondences. For instance, for a pair images with $1600\times 1200$px resolution, where correspondences are computed using a dense feature map with a resolution of $100\times 75$, the output correspondences are localised within an error of $8$ pixels. This can be problematic if correspondences are used for tasks such as pose estimation, where small errors in the localisation of correspondences in image-space can yield high camera pose errors in 3D space.

In this paper, we devise strategies to overcome the limitations of the original NCNet method, while keeping its main advantages, such as the usage of dense feature maps which avoids the issue of missing detections, and the processing of multiple matching hypotheses to avoid early matching errors. Our efficient Sparse-NCNet approach is described next.

\begin{figure}
    \centering
    \includegraphics[width=\textwidth]{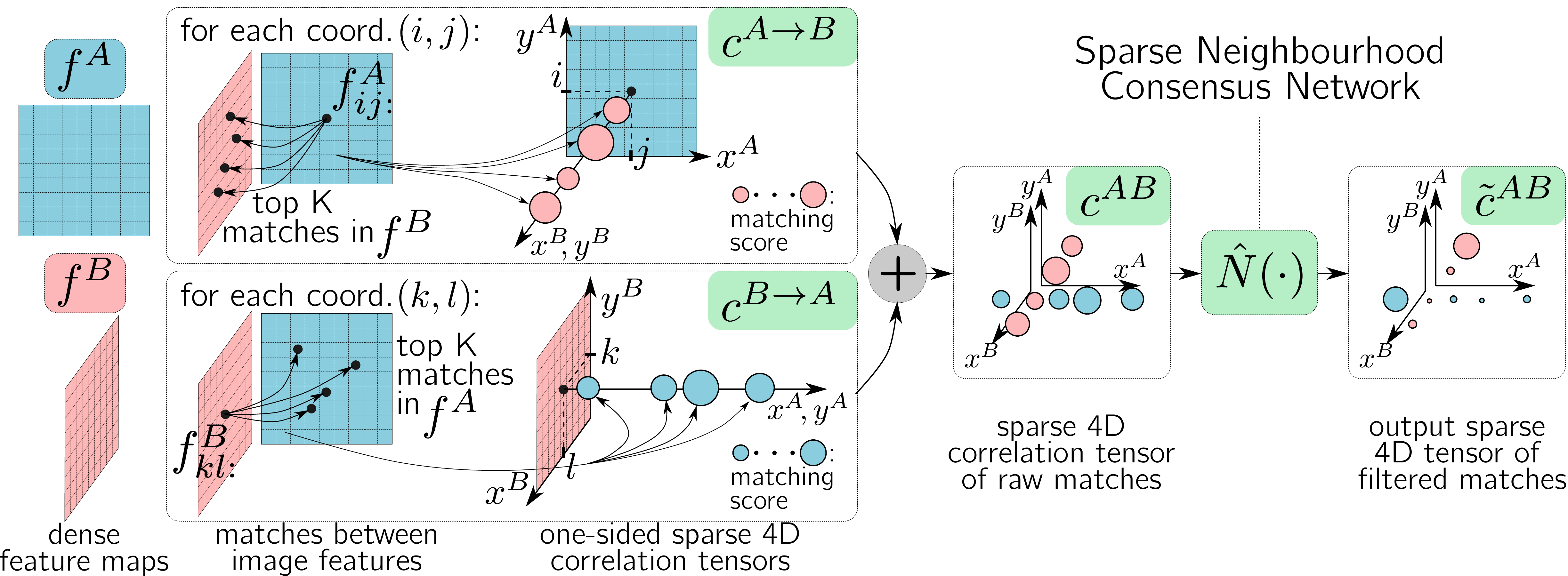}
    \caption{\textbf{Overview of Sparse-NCNet.} From the dense feature maps $f^A$ and $f^B$, their top K matches are computed and stored in the one-sided sparse 4D correlation tensors $c^{A\rightarrow B}$ and $c^{B\rightarrow A}$, which are later combined to obtain the symmetric sparse correlation tensor $c^{AB}$. The raw matching score values in $c^{AB}$ are processed by the 4D Sparse-NCNet $\hat{N}(\cdot)$ producing the output tensor $\tilde{c}^{AB}$ of filtered matching scores.}
    \label{fig:model}
\end{figure}

\subsection{Sparse-NCNet: Efficient Neighbourhood Consensus Networks}
In this section, we describe the Sparse-NCNet approach in detail. An overview is presented in Fig.~\ref{fig:model}. Similar to NCNet, the first stage of our proposed method consists in dense feature extraction. Given a pair of RGB input images  $(I^A, I^B)\in \mathbb{R}^{H\times W\times 3}$, $L2$-normalized dense features $(f^A, f^B)\in \mathbb{R}^{h\times w\times c}$ are extracted via a fully convolutional network $F(\cdot)$: 

\begin{equation}
    f^A=F(I^A), f^B=F(I^B).
\end{equation} 

Then, these dense features are matched and stored into a \emph{sparse correlation tensor}. Contrary to the original NCNet formulation, where \emph{all} the pairwise matches between the dense features are stored and processed,
we propose to keep \emph{only the top $K$ matches} for a given feature, measured by the cosine similarity. In detail, each feature $f^A_{ij:}$ from image $A$ at position $(i,j)$ is matched with its \emph{K nearest-neighbours} in $f^B$, and vice versa. The one-sided sparse correlation tensor, matching from image $A$ to image $B$ ($A\rightarrow B$) is then described as:
\begin{equation}
    c^{A\rightarrow B}_{ijkl}=
  \begin{cases}
    \langle f^A_{ij:},f^B_{kl:}\rangle & \text{if } f^B_{kl:} \text{ within K-NN of } f^A_{ij:} \\ 0 & \text{otherwise}
  \end{cases}.
\end{equation}

To make the sparse correlation map invariant to the ordering of the input images, we also perform this in the reverse direction ($B\rightarrow A$),  and add the two one-sided correlation tensors together to obtain the final (symmetric) \emph{sparse correlation tensor}:
\begin{equation}
    c^{AB}=c^{A\rightarrow B} + c^{B\rightarrow A}. 
    \label{eq:cAB}
\end{equation}
\noindent This tensor uses a sparse representation, where only non-zero elements need to be stored. Note that the number of stored elements is, at most, $h\times w \times K \times 2$ which is in practice much less than  the $h\times w \times h \times w$ elements of the dense correlation tensor, obtaining great memory savings in both the storage of this tensor and its subsequent processing. For example, for a feature map of size $100\times 75$ and $K=10$, the sparse representation takes $3.43$MB
vs.\ $215$MB of the dense representation, resulting in a $12\times$ reduction of the processing time. In the case of feature maps with $200\times 150$ resolution, the sparse representation takes $13.7$MB vs.\ $3433$MB for the dense representation. This allows Sparse-NCNet to also process feature maps at this resolution, something that was not possible with NCNet due to the high memory requirements.
The proposed \emph{sparse correlation tensor} is a compromise between the common procedure of taking the best scoring match and the approach taken by NCNet, where all pairwise matches are stored. In this way, we can keep sufficient information in order avoid early mistakes, while keeping low memory consumption and processing time. 

Then the sparse correlation tensor is processed by a permutation-invariant CNN ($\hat{N}(\cdot)$), to produce the output filtered correlation map $\tilde{c}^{AB}$:
\begin{equation}
    \tilde{c}^{AB} = \hat{N}(c^{AB}).
    \label{eq:ctilde_sparse}
\end{equation}
\noindent The permutation invariant CNN $\hat{N}(\cdot)$ consists of applying the 4D CNN $N(\cdot)$ twice such that the same output matches are obtained regardless of the order of the input images:
\begin{equation}
    \hat{N}(c^{AB}) = N(c^{AB}) + \big(N\big((c^{AB})^T\big)\big)^T,
    \label{eq:nprime}
\end{equation}

\noindent where by transposition we mean exchanging the first two dimensions with the last two dimensions, which correspond to the coordinates of the two input images.
The 4D CNN $N(\cdot)$ operates on the 4D space of correspondences, and is trained to perform the neighbourhood consensus filtering. Note that while $N(\cdot)$ is a sparse CNN using  submanifold sparse convolutions~\cite{3DSemanticSegmentationWithSubmanifoldSparseConvNet}, where the active sites between the sparse input and output remain constant, the convolution kernel filters are dense (\ie hypercubic).

While in the original NCNet method, a soft mutual nearest-neighbour operation $M(\cdot)$ is also performed, we have removed it as we noticed its effect was not significant when operating on the sparse correlation tensor.
From the output correlation tensor $\tilde{c}^{AB}$, the output matches are computed by applying argmax at each coordinate:
\begin{equation}
    \big((i,j),(k,l)\big) \text{ a match if } \begin{cases}(i,j) = \underset{(a,b)}{\argmax} \ \tilde{c}^{AB}_{abkl},\text{ or} \\
    (k,l) = \underset{(c,d)}{\argmax} \ \tilde{c}^{AB}_{ijcd}
    \end{cases},
    \label{eq:matches}
\end{equation}

\noindent where $(i,j)$ is the match coordinate in the sampling grid of $f^A$, and $(k,l)$ is the match coordinate in the sampling grid of $f^B$.

\subsection{Match relocalisation by guided search}
While the sparsification of the correlation tensor presented in the previous section allows processing higher resolution feature maps, these are still several times smaller in resolution than the input images. Hence, they are not suitable for applications that require (sub)pixel feature localisation such as camera pose estimation or 3D-reconstruction.

To address this issue, in this paper we propose a two-stage relocalisation module based on the idea of guided search. The intuition is that we search for accurately localised matches on $2h\times 2w$ resolution dense feature maps, guided by the coarse matches output by Sparse-NCNet at $h\times w$ resolution.
\begin{figure}[t]
    \centering
    \includegraphics[width=\textwidth]{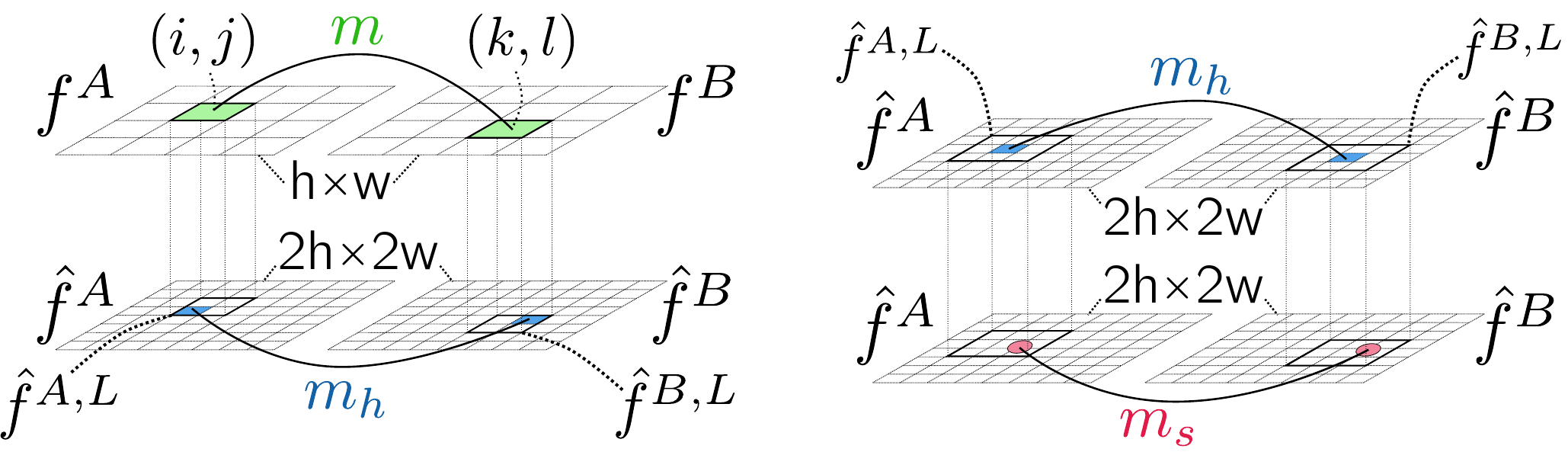}
    \begin{subfigure}{0.43\textwidth}
    \caption{Hard relocalisation}
    \end{subfigure}
    \hfill
    \begin{subfigure}{0.43\textwidth}
    \caption{Soft relocalisation\ \ \ \ \ \ \ \ \ }
    \end{subfigure}
    \vspace*{-1em}
    \caption{\textbf{Two-stage relocalisation module.} (a) The hard relocalisation step allows to increase by $2\times$ the localisation accuracy of the matches $m$ outputted by Sparse-NCNet, which are defined on the $h\times w$ feature maps $f^A$ and $f^B$. This is done by keeping the most similar match $m_h$ between two $2\times 2$ local features $\hat{f}^{A,L}$ and $\hat{f}^{B,L}$, cropped from the $2h\times 2w$ feature maps $\hat{f}^A$ and $\hat{f}^B$. (b) The soft relocalisation step then refines the position of these matches in the $2h\times 2w$ grid, by computing sub-feature-grid soft localisation displacements based on the softargmax operation.}
    \label{fig:reloc}
\end{figure}
For this, dense features are first extracted at twice the normal resolution $(\hat{f}^{A},\hat{f}^{B})\in \mathbb{R}^{2h\times 2w \times c}$, which is done by upsampling the input image by $2\times$ before feeding it into the feature extraction CNN $F(\cdot)$. Note that these higher resolution features are used for relocalisation only, \ie they are not used to compute the correlation tensor or processed by the 4D CNN for match-filtering, which would be too expensive. Then, these dense features are downsampled back to the normal $h\times w$ resolution by applying a $2\times2$ max-pooling operation with a stride of $2$, obtaining $f^A$ and $f^B$. These low resolution features $(f^A,f^B)\in \mathbb{R}^{h\times w \times c}$ are processed by Sparse-NCNet, which outputs matches in the form $m=\big((i,j),(k,l)\big)$, with the coordinates $(i,j)$ and $(k,l)$ indicating the position of the match in $f^A$ and $f^B$, respectively, as described by~\eqref{eq:matches}.

Having obtained the output matches in $h\times w$ resolution, the first step (hard relocalisation) consists in finding the best equivalent match in the $2h\times 2w$ resolution grid. This is done by analysing the matches between two local crops of the high resolution features $\hat{f}^A$ and $\hat{f}^B$, and keeping the highest-scoring one. The second step (soft relocalisation) then refines this correspondence further, by obtaining a sub-feature accuracy in the $2h\times 2w$ grid. These two relocalisation steps are illustrated in Fig.~\ref{fig:reloc}, and are now described in detail.

\paragraph{Hard relocalisation.}
The first step is hard relocalisation, which can improve localisation accuracy by $2\times$. For each match $m=\big((i,j),(k,l)\big)$,  the $2\times$ upsampled coordinates $\big((2i,2j),(2k,2l)\big)$ are first computed, and $2\times 2$ local feature crops $\hat{f}^{A,L},\hat{f}^{B,L}\in \mathbb{R}^{2\times 2\times c}$ are sampled around these coordinates from the high resolution feature maps $\hat{f}^A$ and $\hat{f}^B$:
\begin{equation}
    \hat{f}^{A,L} = (\hat{f}^{A}_{ab:})_{\substack{2i \le a \le 2i+1\\ 2j \le b \le 2j+1}},
\end{equation}
\noindent and similarly for $\hat{f}^{B,L}$. This is done using a ROI-pooling operation~\cite{girshick2015fast}. Finally, exhaustive matches between the local feature crops $\hat{f}^{A,L}$ and $\hat{f}^{B,L}$ are computed, and the output of the hard relocalisation module is the displacement associated with the maximal matching score:
\begin{equation}
    \Delta m_{h} = \big((\delta i,\delta j),(\delta k,\delta l)\big)  = \underset{(a,b),(c,d)}{\argmax} 
    \langle \hat{f}^{A,L}_{ab:},\hat{f}^{B,L}_{cd:} \rangle.
\end{equation}
\noindent Then, the final match location from the hard relocalisation stage is computed as:
\begin{equation}
    m_h = 2m + \Delta m_h  = \big((2i+\delta i,2j+\delta j),(2k+\delta k,2l+\delta l)\big).
\end{equation}
Note that the relocalised matches $m_h$ are defined in a $2h\times 2w$ grid, therefore obtaining a $2\times$ increase in localisation accuracy with respect to the initial matches $m$, which are defined in a $h\times w$ grid. Also note that while the implementation is different, the effect of the proposed hard relocalisation is similar to the max-argmax operation used in NCNet~\cite{NCNet}, while being more memory efficient as it avoids the computation of the a dense correlation tensor in high resolution.

\paragraph{Soft relocalisation.}
The second step consists of a soft relocalisation operation that  obtains sub-feature localisation accuracy in the $2h\times 2w$ grid of high resolution features $\hat{f}^A$ and $\hat{f}^B$. For this, new $3\times 3$ local feature crops $(\hat{f}^{A,L},\hat{f}^{B,L})\in \mathbb{R}^{3\times 3\times c}$ are sampled around the coordinates of the estimated matches $m_h$ from the previous relocalisation stage. Note that no upsampling of the coordinates is done in this case, as the matches are already in the $2h\times 2w$ range. Then, soft relocalisation displacements are computed by performing the softargmax operation~\cite{yi2016lift} on the matching scores between the central feature of $\hat{f}^{A,L}$ and the whole of $\hat{f}^{B,L}$, and vice versa:
\begin{equation}
    \Delta m_s=\big((\delta i,\delta j),(\delta k,\delta l)\big)\text{ where } \begin{cases} (\delta i,\delta j) = \underset{(a,b)}{\softargmax} \,\langle \hat{f}^{A,L}_{ab:},\hat{f}^{B,L}_{11:} \rangle \\
    (\delta k,\delta l) = \underset{(c,d)}{\softargmax} \,\langle \hat{f}^{A,L}_{11:},\hat{f}^{B,L}_{cd:} \rangle
    \end{cases}\hspace{-0.5em}.
    \label{eq:softreloc}
\end{equation}
The intuition of the softargmax operation is that it computes a weighted average of the candidate positions in the crop where the weights are given by the softmax of the matching scores. The final matches from soft relocalisation are obtained by applying the soft displacements to the matches from hard relocalisation: $m_s = m_h + \Delta m_s$.

\section{Experimental evaluation}
We evaluate the proposed Sparse-NCNet method on three different benchmarks: (i) HPatches Sequences, which evaluates the matching task directly, (ii) InLoc, which targets the problem of indoor 6-dof camera localisation and (iii) Aachen Day-Night, which targets the problem of outdoor 6-dof camera localisation with challenging day-night illumination changes. We first present the implementation details followed by the results on these three benchmarks.

\paragraph{Implementation details.} 
We train the Sparse-NCNet model following the training protocol from~\cite{NCNet}. We use the IVD dataset with the weakly-supervised mean matching score loss for training~\cite{NCNet}. The 4D CNN $N(\cdot)$ has two sparse convolution layers with $3^4$ sized kernels, with $16$ output channels in the hidden layer. A value of $K=10$ is used for computing $c^{AB}$~\eqref{eq:cAB}. The model is implemented using PyTorch~\cite{pytorch}, MinkowskiEngine~\cite{choy20194d} and Faiss~\cite{faiss}, and trained for 5 epochs using Adam~\cite{kingma2015adam} with a learning rate of $5\times 10^{-4}$. A pretrained ResNet-101 (up to \texttt{conv\_4\_23}) with no strided convolutions in the last block is used as the feature extractor $F(\cdot)$. This feature extraction model is not finetuned as the training dataset is small (3861 image pairs) and that would lead to overfitting and loss of generalisation. The softargmax operation in~\eqref{eq:softreloc} uses a temperature value of 10.

\subsection{HPatches Sequences}
The HPatches Sequences~\cite{hpatches2017cvpr} benchmark assesses the matching accuracy under strong \emph{viewpoint} and \emph{illumination} variations. We follow the evaluation procedure from~\cite{Dusmanu2019CVPR}, where 108 image sequences are employed, each from a different planar scene, and each containing 6 images. The first image from each sequence is matched against the remaining 5 images. The benchmark employs 56 sequences with viewpoint changes, and constant illumination conditions, and 52 sequences with illumination changes and constant viewpoint. The metric used for evaluation is the mean matching accuracy (MMA)~\cite{Dusmanu2019CVPR}. Further details about this metric are provided in Appendix~\ref{sec:hp_qual}.

\begin{figure}[t]
\centering
\begin{subfigure}[t]{\textwidth}

\newcommand{\ra}[1]{\renewcommand{\arraystretch}{#1}}
\setlength{\tabcolsep}{3.5pt}
\ra{1.1}
{\fontsize{8}{9}\selectfont 
\begin{tabular}{@{}l@{}lccccc@{}}\toprule

& \textbf{Method}
&  \begin{tabular}[c]{@{}c@{}}\textbf{Feature}\\ \textbf{resolution}\end{tabular} & \begin{tabular}[c]{@{}c@{}}\textbf{Reloc.}\\ \textbf{method}\end{tabular} & \begin{tabular}[c]{@{}c@{}}\textbf{Reloc.}\\ \textbf{resolution}\end{tabular} & \begin{tabular}[c]{@{}c@{}}\textbf{Mean}\\ \textbf{time (s)}\end{tabular} & \begin{tabular}[c]{@{}c@{}}\textbf{Peak}\\ \textbf{VRAM (MB)}\end{tabular} \\ \midrule

    A1.\plotlabel{cb3}{triangle*}{dashed} & Sparse-NCNet & $100\times 75$  & --- & ---            &  0.83 & 251  \\
    A2.\plotlabel{cb14}{*}{dashed} & NCNet        & $100\times 75$  & --- & ---            &  9.81 & 5763 \\ \midrule
    B1.\plotlabel{cb7}{|}{dash pattern={on 1.5pt off 2pt on 4pt off 2pt}} & Sparse-NCNet & $100\times 75$  & H   & $200\times150$ &  1.55 & 1164 \\
    B2.\plotlabel{cb10}{square*}{dash pattern={on 1.5pt off 2pt on 4pt off 2pt}} & NCNet        & $100\times 75$  & H   & $200\times150$ & 10.56 & 7580 \\ \midrule
    C1.\plotlabel{cb4}{x}{} & Sparse-NCNet & $100\times 75$  & H+S & $200\times150$ &  1.56 & 1164 \\ 
    C2.\plotlabel{cb1}{}{} & Sparse-NCNet & $200\times 150$ & H+S & $400\times300$ &  7.51 & 2391 \\ \bottomrule
\end{tabular}}
\caption{Time and GPU memory comparison (Tesla T4 GPU)}
\label{fig:hseq_variants_table}

\end{subfigure}

\begin{subfigure}[t]{\textwidth}
\centering
\includegraphics[width=\textwidth]{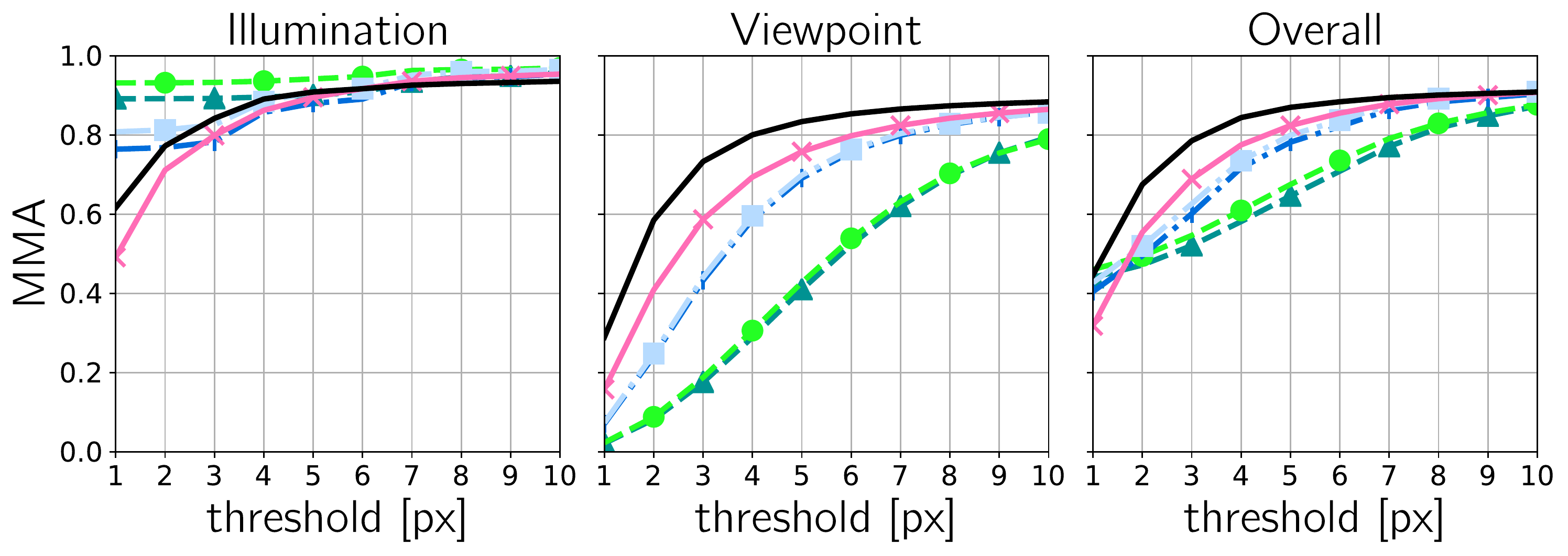}
\caption{MMA on HPatches Sequences}
\label{fig:hseq_variants_plot}
\end{subfigure}
\caption{\textbf{Sparse-NCNet vs.\ NCNet on HPatches}. 
Sparse-NCNet can obtain equivalent results to NCNet, both without relocalisation (\cf A1 vs.\ A2), and with hard relocalisation (H) (\cf B1 vs.\ B2), while greatly reducing execution time and memory consumption. The proposed two-stage relocalisation (H+S) brings an improvement in matching accuracy with a minor increase in execution time (\cf C1 vs.\ B1). Finally, the reduced memory consumption in Sparse-NCNet allows for processing in higher resolution, which produces the best results, while still being faster and more memory efficient than NCNet (\cf C2 vs.\ B2).\label{fig:hseq_variants}}
\vspace{-0.5cm}
\end{figure}

\paragraph{Sparse-NCNet vs.\ NCNet.} In Fig.~\ref{fig:hseq_variants} we compare the matching quality of the proposed Sparse-NCNet model and the NCNet model. We first compare both methods under equal conditions, both without relocalisation (methods A1 vs.\ A2), and with hard relocalisation only (methods B1 vs.\ B2). The results in Fig.~\ref{fig:hseq_variants} show that Sparse-NCNet can obtain significant reductions in processing time and memory consumption, while keeping almost the same matching performance. In addition, our proposed two-stage relocalisation module can improve performance with a minor increase in processing time (methods C1 vs.\ B1). Finally, the reduced memory consumption allows for processing of higher resolution $200\times 150$ feature maps, which is not possible for NCNet. Our proposed method in higher resolution (method C2) produces the best results while still being 30\% faster and $3\times$ more memory efficient than the best NCNet variant (method B2).

\begin{figure}[t]
    \centering
    \includegraphics[width=\textwidth]{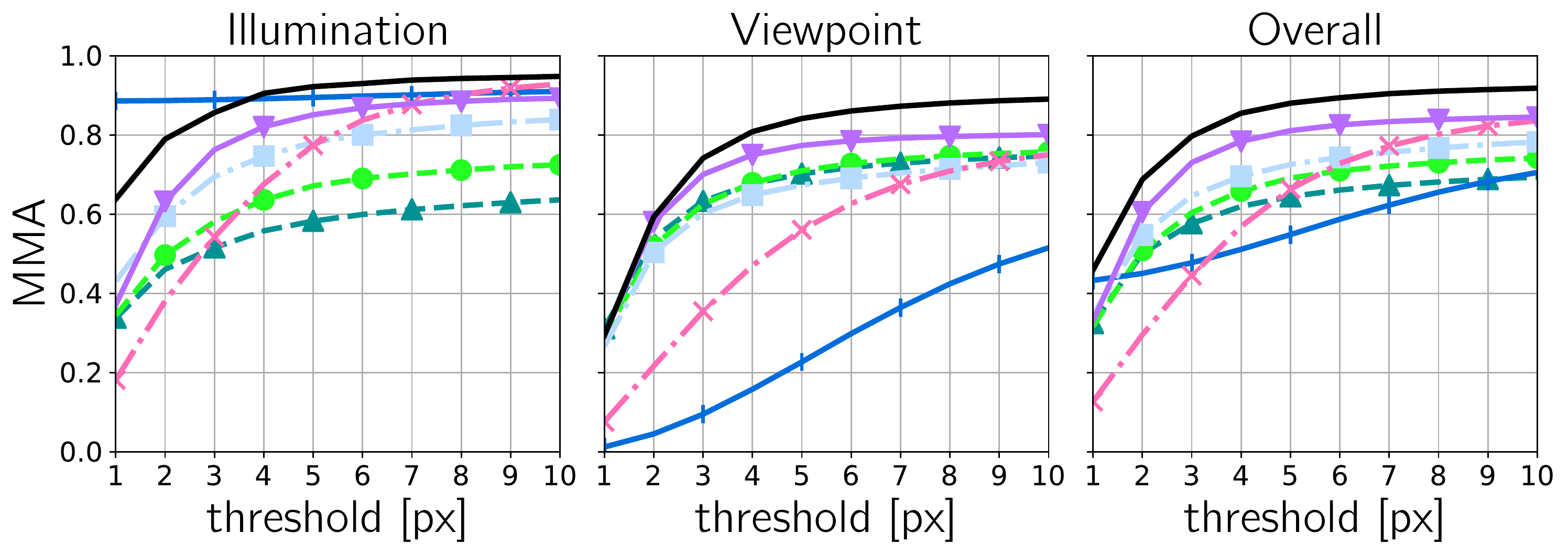}
    {\fontfamily{cmss}\fontsize{8}{10}\selectfont
    \begin{tabular}{cl}
     \plotlabelv{cb1}{}{} & Sparse-NCNet
     \end{tabular}
     \begin{tabular}{cl}
     \plotlabelv{cb8}{triangle*,mark options={rotate=180}}{} & R2D2~\cite{r2d2}
     \end{tabular}
     \begin{tabular}{cl}
     \plotlabelv{cb4}{x}{dash pattern={on 1.5pt off 2pt on 4pt off 2pt}} & D2-Net~\cite{Dusmanu2019CVPR}
     \end{tabular}
     \begin{tabular}{cl}
     \plotlabelv{cb10}{square*}{dash pattern={on 1.5pt off 2pt on 4pt off 2pt}} & SuperPoint~\cite{Detone2018CVPRW}
     \end{tabular}
     \begin{tabular}{cl}
     \plotlabelv{cb7}{|}{} & DELF~\cite{noh2017DELF}
     \end{tabular}
     \begin{tabular}{cl}
     \plotlabelv{cb14}{*}{dashed} & HessAffNet + HN$_{++}$~\cite{HardNet,Mishkin2018AffNet}
     \end{tabular}
     \begin{tabular}{cl}
     \plotlabelv{cb3}{triangle*}{dashed} & Affine Det. + RootSIFT~\cite{mikolajczyk2002affine,arandjelovic2012three} 
     \end{tabular}}
    \vspace{-2mm}
    \caption{\textbf{Sparse-NCNet vs.\ state-of-the-art on HPatches.} The MMA of Sparse-NCNet and several state-of-the-art methods is shown. Sparse-NCNet obtains the best results overall with a large margin over the recent R2D2 method.\label{fig:hseq_baselines}}
    \vspace{-2mm}
\end{figure}

\paragraph{Sparse-NCNet vs.\ state-of-the-art methods.} In addition, we compare the performance of Sparse-NCNet against several methods, including state-of-the-art trainable methods such as SuperPoint~\cite{Detone2018CVPRW}, D2-Net~\cite{Dusmanu2019CVPR} or R2D2~\cite{r2d2}. The mean-matching accuracy results are presented in Fig.~\ref{fig:hseq_baselines}. For all other methods, the top 2000 features points where selected from each image, and matched enforcing mutual nearest-neighbours, yielding approximately 1000 correspondences per image pair. For Sparse-NCNet, the top 1000 correspondences where selected for each image pair, for a fair comparison. Sparse-NCNet obtains the best results for the \emph{illumination} sequences for thresholds higher than $4$ pixels, and in the \emph{viewpoint} sequences for all threshold values. Sparse-NCNet obtains the best results overall, with a large margin over the state-of-the-art R2D2 method. We believe this could be attributed to the usage of dense descriptors (which avoid the loss of detections) together with an increased matching robustness from performing neighbourhood consensus. Qualitative examples and comparison with other methods are presented in Appendix~\ref{sec:hp_qual}.

\begin{figure}[t]
    \centering
    \begin{subfigure}{0.44\textwidth}
    \centering
    \includegraphics[width=\textwidth]{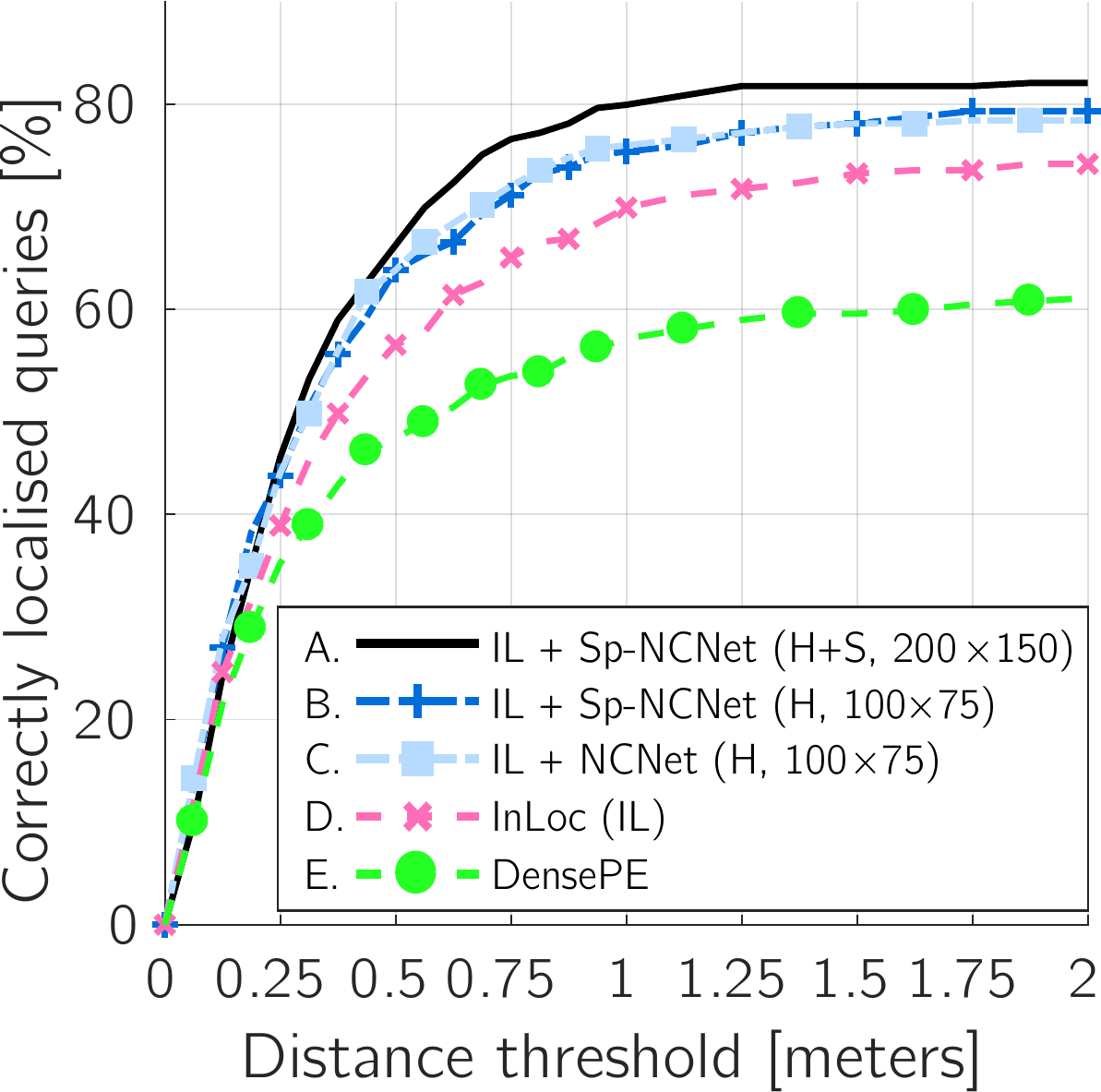}
    \end{subfigure}
    ~
    \begin{subfigure}{0.44\textwidth}
    \centering
    \includegraphics[width=0.45\textwidth]{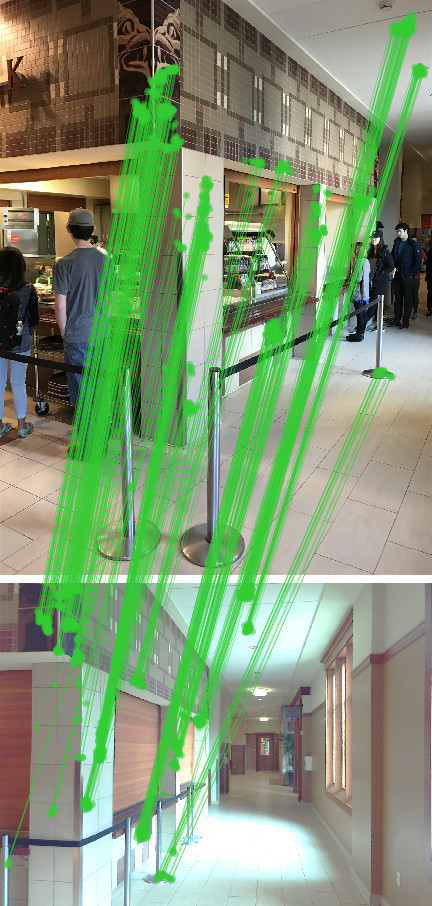}
    \includegraphics[width=0.45\textwidth]{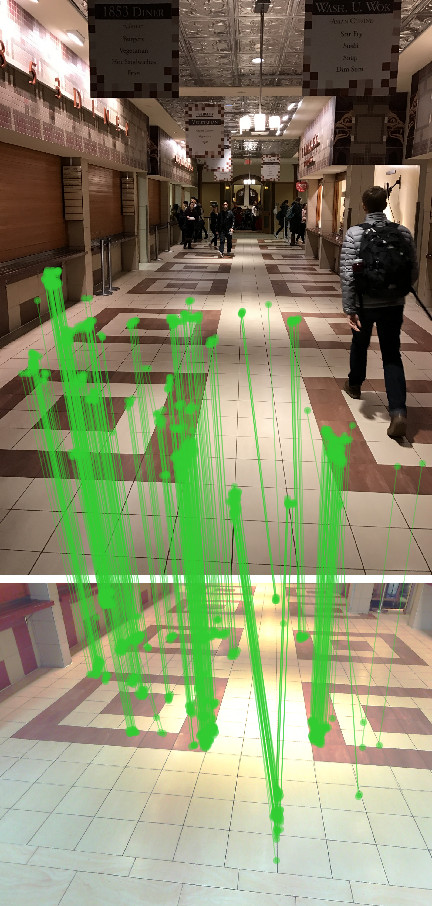}
    \end{subfigure}
    \vspace{-1mm}
    \caption{\textbf{Results on the InLoc benchmark for long-term indoor localization.} 
    (Left) Our proposed method (A) obtains state-of-the-art results on this benchmark. (Right) Our method  obtains correspondences in challenging indoor scenes with repetitive patterns and low amount of texture. Top: query images. Bottom: matched database images captured from different viewpoints. Correspondences produced by our approach are overlaid in green. The query and database images were taken several months apart.\label{fig:inloc}}
    \vspace{-3mm}
\end{figure}

\subsection{InLoc benchmark}
The InLoc benchmark~\cite{Taira18} targets the problem of indoor localisation. It contains a set of \emph{database} images of a building, obtained with a 3D scanner, and a set of \emph{query} images from the same building, captured with a cell-phone several months later. The task is then to obtain the 6-dof camera positions of the query images. We follow the DensePE approach proposed~\cite{Taira18} to find the top 10 candidate database images for each query, and employ Sparse-NCNet to obtain matches between them. Then, we follow again the procedure in~\cite{Taira18} to obtain the final estimated 6-dof query pose, which consists of running PnP~\cite{gao2003complete} followed by dense pose verification~\cite{Taira18}. 

The results are presented in Fig.~\ref{fig:inloc}. First, we observe that Sparse-NCNet with hard relocalisation (H) and a resolution of $100\times 75$ obtains equivalent results to NCNet (methods B vs.\ C), while being almost $7\times$ faster and requiring $6.5\times$ less memory, confirming what was already observed in the HPatches benchmark (\cf B1 vs.\ B2 in Fig.~\ref{fig:hseq_variants_table}). Moreover, our proposed Sparse-NCNet method with two-stage relocalisation (H+S) in the higher $200\times 150$ resolution (method A) obtains the best results and sets a new state-of-the-art for this benchmark. Recall that it is impossible to use the original NCNet on the higher resolution due to its excessive memory requirements. More qualitative examples are included in Appendix~\ref{sec:inloc_qual}.

\begin{table}[t]
    \vspace{-0.3cm} 
    \captionof{table}{\textbf{Results on Aachen Day-Night.}
    Sparse-NCNet is able to localise a similar number of queries than R2D2 and D2-Net.}
    \centering
    {\fontsize{8}{9}\selectfont 
    \begin{tabular}{@{}l@{\hspace{1mm}} c@{\hspace{6mm}}c@{\hspace{2mm}}c@{}}
    \toprule 
    & \multicolumn{3}{l}{\textbf{Correctly localised queries (\%)}} \\
    \textbf{Method} & $0.5$m, $2^\circ$ & $1.0$m, $5^\circ$ & $5.0$m, $10^\circ$  \\
    \midrule
    RootSIFT~\cite{lowe2004distinctive,arandjelovic2012three} & $36.7$ & $54.1$ & $72.5$  \\
    DenseSfM~\cite{sattler2018benchmarking} &  $39.8$ & $60.2$ & $84.7$ \\
    HessAffNet\,+\,HN$_{++}$~\cite{HardNet,Mishkin2018AffNet} & $39.8$ & $61.2$ & $77.6$ \\
    DELF~\cite{noh2017DELF} & $38.8$ & $62.2$ & $85.7$ \\ 
    SuperPoint~\cite{Detone2018CVPRW} & $42.8$ & $57.1$ & $75.5$ \\
    D2-Net~\cite{Dusmanu2019CVPR} &  $44.9$ & $66.3$ & $\mathbf{88.8}$  \\
    D2-Net (Multi-scale)~\cite{Dusmanu2019CVPR} &  $44.9$ & $64.3$ & $\mathbf{88.8}$  \\
    R2D2 ($\text{patch}=16$)~\cite{r2d2} & $44.9$ & $67.3$ & $87.8$ \\
    R2D2 ($\text{patch}=8$)~\cite{r2d2} & $\mathbf{45.9}$ & $66.3$ & $\mathbf{88.8}$ \\
    Sparse-NCNet (H, $200\times 150$) & $44.9$ & $\mathbf{68.4}$ & $86.7$ \\
    \bottomrule 
    \end{tabular}}
    \vspace{-0.2cm}
    \label{tab:aachen}
\end{table}

\subsection{Aachen Day-Night}
The Aachen Day-Night benchmark~\cite{sattler2018benchmarking} targets 6-dof outdoor camera localisation under challenging illumination conditions. It contains 98 night-time query images from the city of Aachen, and a shortlist of 20 day-time images for each night-time query. Sparse-NCNet is used to obtain matches between the query and images in the short-list. The resulting matches are then processed by the 3D reconstruction software COLMAP~\cite{schoenberger2016sfm} to obtain the estimated query poses. 

The results are presented in Table~\ref{tab:aachen}. Sparse-NCNet presents a similar performance to the state-of-the-art methods D2-Net~\cite{Dusmanu2019CVPR} and R2D2~\cite{r2d2}. Note that the results of these three different methods differ by only a few percent, which represents only 1 or 2 additionally localised queries, from the 98 total night-time queries. The proposed Sparse-NCNet obtains state-of-the-art results for the $1$m and $5^\circ$ threshold, being able to localise $68.4\%$ of the queries (67 out of 98). One qualitative example from this benchmark is presented in Fig.~\ref{fig:teaser}, and more are included in Appendix~\ref{sec:aachen_qual}.

\section{Conclusion}
In this paper we have developed Sparse Neighbourhood Consensus Networks for efficiently estimating correspondences between images.
Our approach overcomes the main limitations of the original Neighbourhood Consensus Networks that demonstrated promising results on challenging matching problems, making these models practical and widely applicable. 
The proposed model jointly performs feature extraction, matching and robust match filtering in a computationally efficient manner, outperforming state-of-the-art results on two challenging matching benchmarks. The entire pipeline is end-to-end trainable, which opens-up the possibility for including additional modules for specific downstream problems such as camera pose estimation or 3D reconstruction.  

\vspace{1em}

{\noindent\small \textbf{Acknowledgements.} This work was partially supported by ERC grant LEAP No. 336845, the European Regional Development Fund under project IMPACT (reg. no. CZ.02.1.01/0.0/0.0/15 003/0000468), Louis Vuitton ENS Chair on Artificial Intelligence, and 
the French government under management of Agence Nationale de la Recherche as part of the ``Investissements d'avenir" program, reference ANR-19-P3IA-0001 (PRAIRIE 3IA Institute).}

\bibliographystyle{splncs04}
\bibliography{shortstrings,egbib}

\clearpage

\appendix

\section*{Appendices}

In this appendices we present insights about the way Sparse-NCNet operates (Appendix~\ref{sec:intuitions}) and additional qualitative results on the HPatches Sequences (Appendix~\ref{sec:hp_qual}), InLoc (Appendix~\ref{sec:inloc_qual}) and Aachen Day-Night (Appendix~\ref{sec:aachen_qual}) benchmarks.

\section{Insights about Sparse-NCNet\label{sec:intuitions}}
\begin{figure}
    \centering
    \vspace*{-1em}
    \setlength{\tabcolsep}{4pt}
    \begin{tabular}{ccc}
    \includegraphics[width=3.6cm]{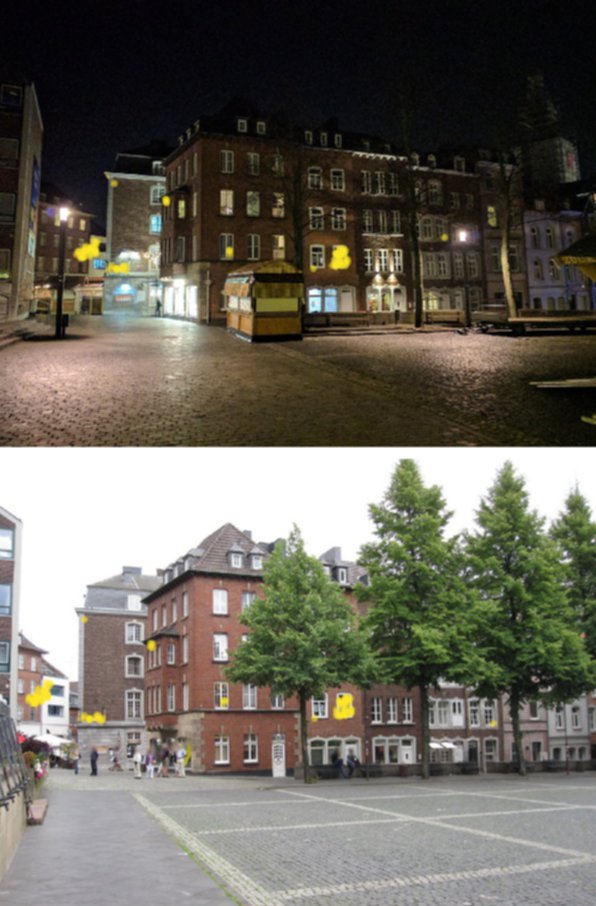} &
    \includegraphics[width=3.6cm]{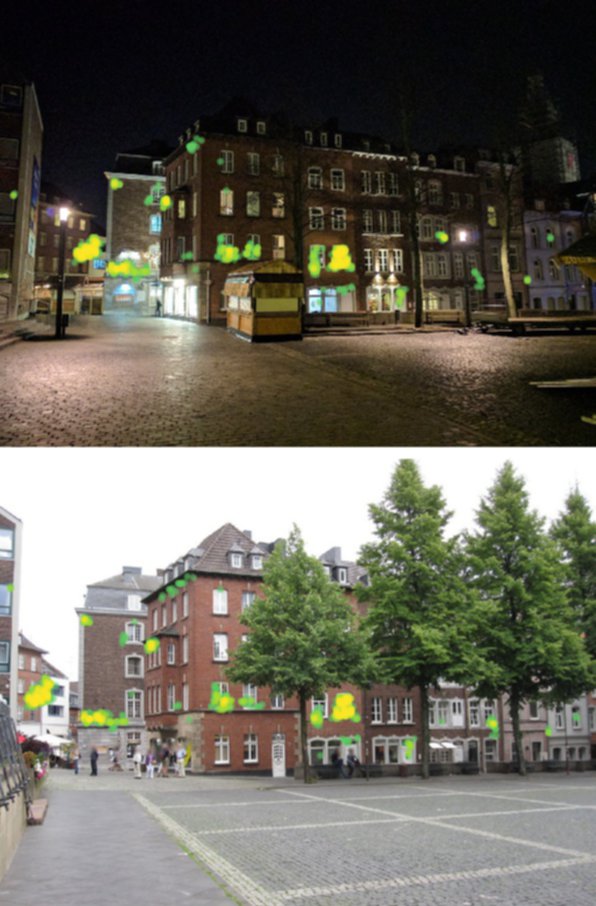} &
    \includegraphics[width=3.6cm]{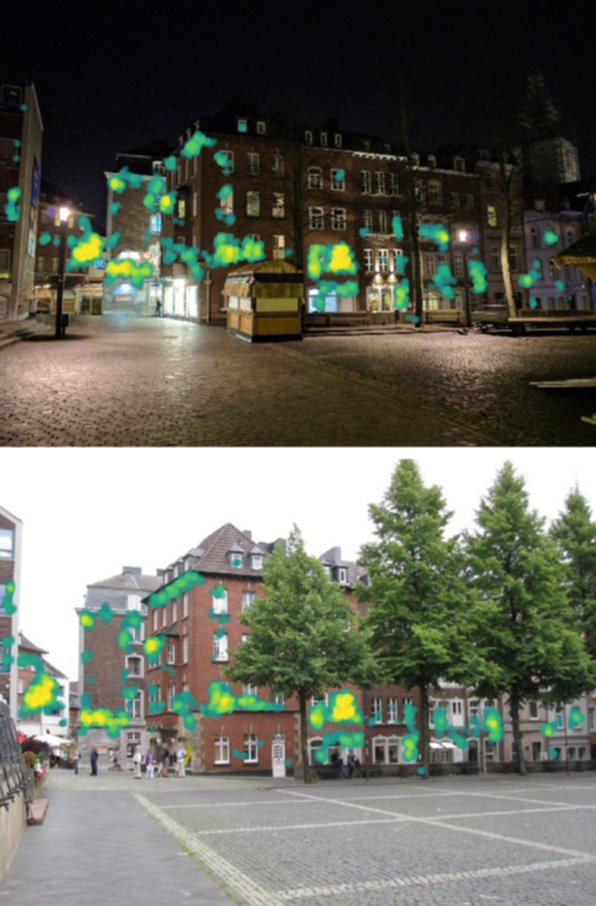} \\[1.5em]
    
    \includegraphics[width=3.6cm]{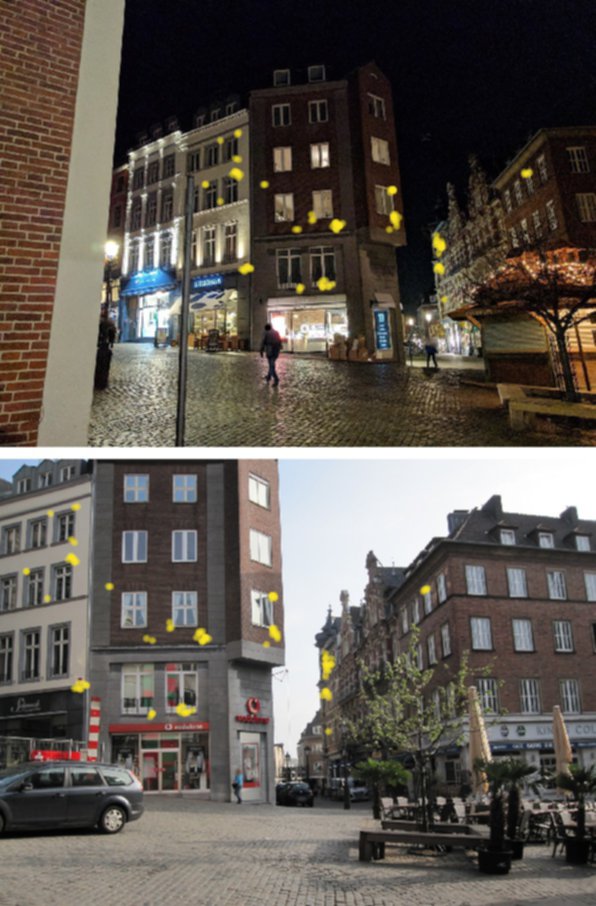} &
    \includegraphics[width=3.6cm]{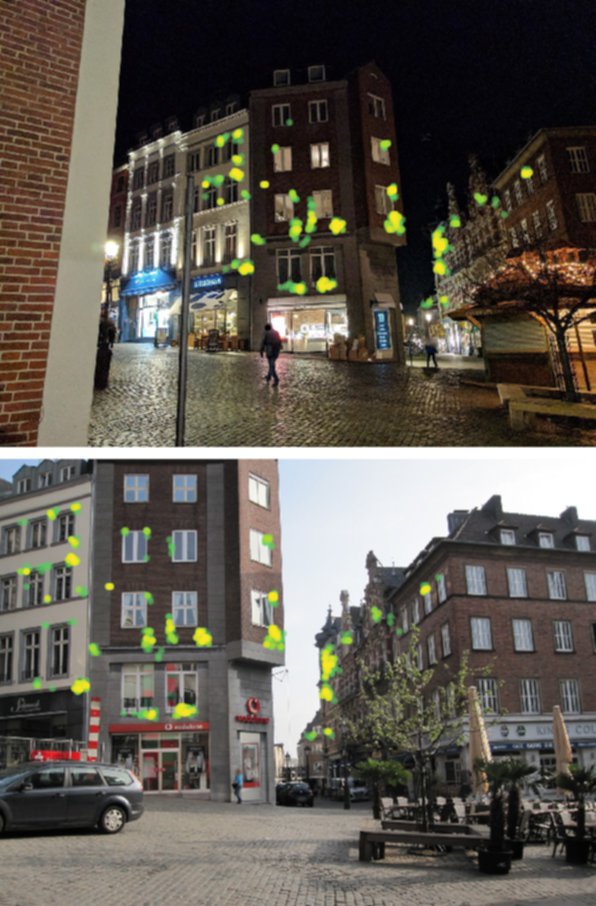} &
    \includegraphics[width=3.6cm]{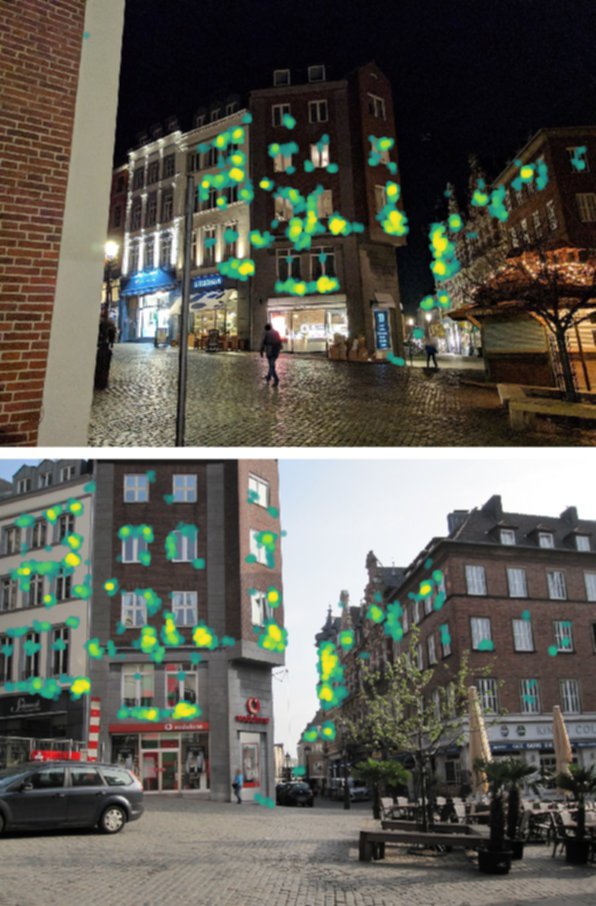} \\[1.5em]
    
    \includegraphics[width=3.6cm]{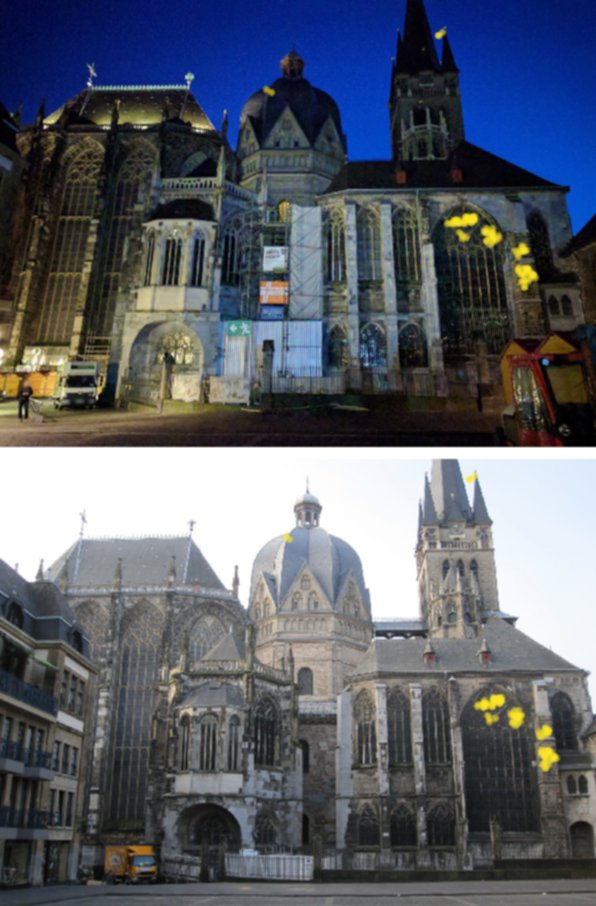} &
    \includegraphics[width=3.6cm]{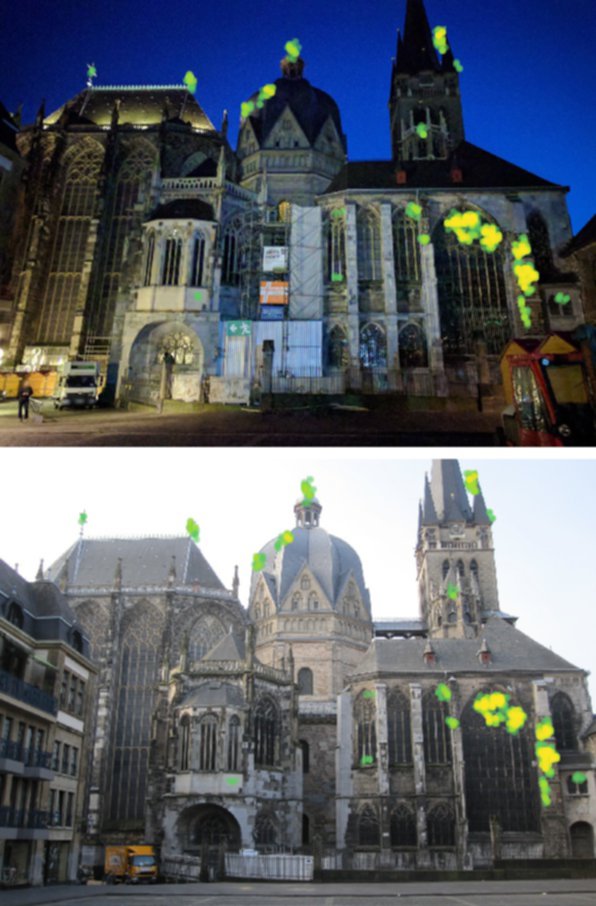} &
    \includegraphics[width=3.6cm]{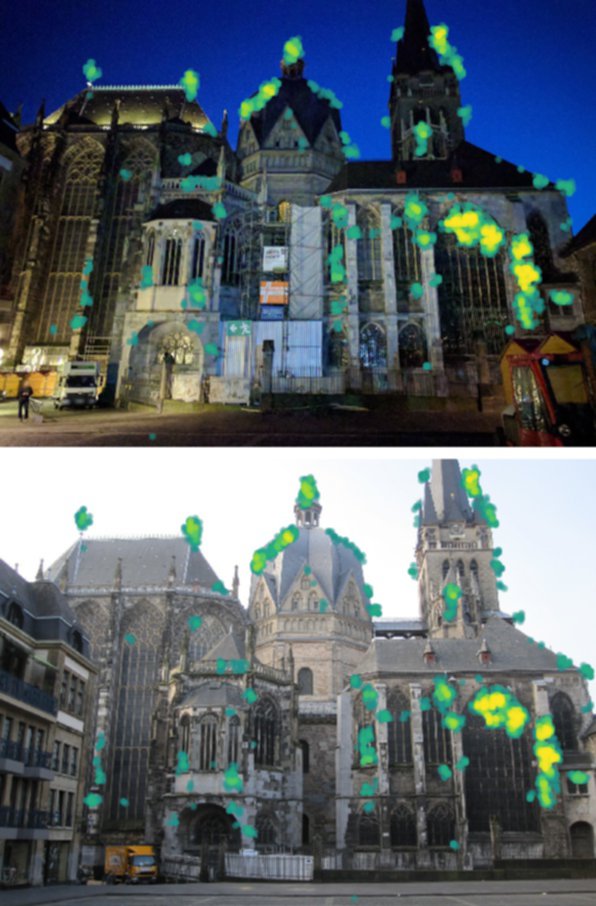} \\
    $N=100$ & $N=400$ & $N=1600$
    \end{tabular}

    \caption{\textbf{Insights about Sparse-NCNet.}  We show the top $N$ matches between each pair of images for different values of $N$. The strength of the match is shown by color (the more yellow the stronger). Please note how new matches tend to appear close to high scoring matches, demonstrating the propagation of information in Sparse-NCNet.}
    \label{fig:insights}
\end{figure}

In this section we provide additional insights about the way Sparse-NCNet operates, which differs from traditional local feature detection and matching methods. In Fig.~\ref{fig:insights} we plot the top $N$ matches produced by Sparse-NCNet for different values of $N$: $100$ (left column), $400$ (middle column) and $1600$ (right column). By comparing the middle column (showing the top 400 matches) with the left column (showing the top 100), we can observe that many of the additional 300 matches are close to the initial 100 matches. A similar effect is observed when comparing the right column (top 1600 matches) with the middle column (top 400 matches). This could be attributed to the fact that Sparse-NCNet  propagates information from the strongest matches to their neighbours. In this sense, strong matches, which are typically non-ambiguous ones, can help in matching their neighbouring features, which might not be so discriminative.

\FloatBarrier

\section{HPatches Sequences benchmark\label{sec:hp_qual}}

\paragraph{Mean matching accuracy.} The Mean Matching Accuracy (MMA) metric is used in the HPatches Sequences benchmark to assess the fraction of correct matches under different tolerance thresholds. It is computed in the following way:

\begin{equation}
    \text{MMA}\big(\{(p^A_i,p^B_i)\}_{i=1}^N; t\big) = \frac{\sum_{i=1}^N \mathlarger{\mathlarger{\mathbbm{1}}}_{\mathsmaller{>0}}\big(t - \| \mathcal{T}_H(p^A_i) - p^B_i \| \big)}{N},
\end{equation}

\noindent where $\{(p^A_i,p^B_i)\}_{i=1}^N$ is the set of matches to be evaluated, $\mathcal{T}_H(p^A_i)$ is the warped point $p^A_i$ using the ground-truth homography $H$, $\mathlarger{\mathlarger{\mathbbm{1}}}_{\mathsmaller{>0}}$ is the indicator function for positive numbers, and $t$ is the chosen tolerance threshold (in pixels).

\begin{figure}
    \centering
    \begin{tabular}{ccccc}
    \smaller SuperPoint~\cite{Detone2018CVPRW} & \smaller D2-Net~\cite{Dusmanu2019CVPR} & \smaller R2D2~\cite{r2d2} & \smaller Sparse-NCNet, 2k & \smaller Sparse-NCNet, 6k\\
    \includegraphics[width=2.3cm]{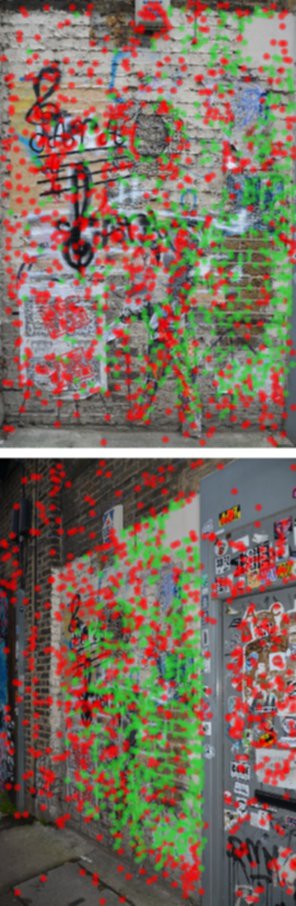} &
    \includegraphics[width=2.3cm]{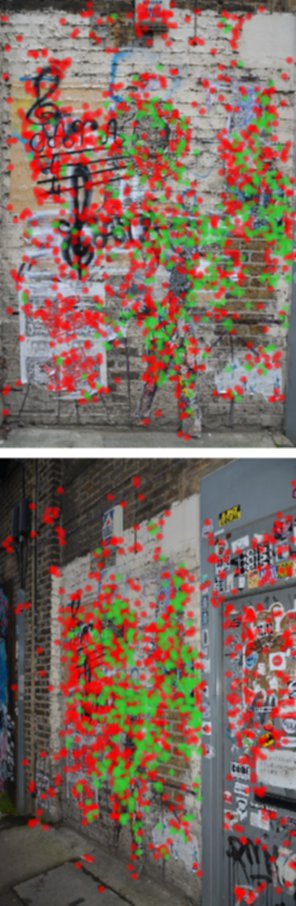} &
    \includegraphics[width=2.3cm]{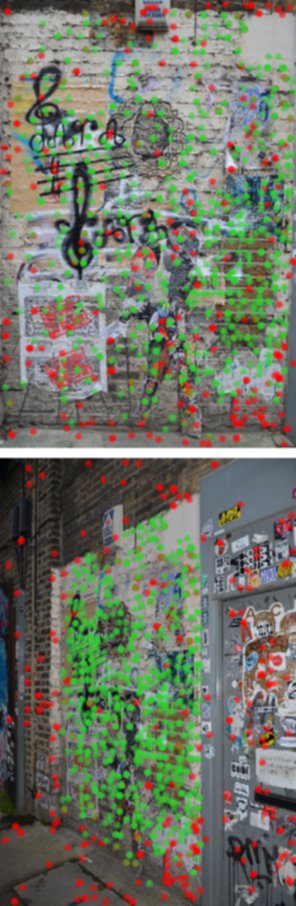} &
    \includegraphics[width=2.3cm]{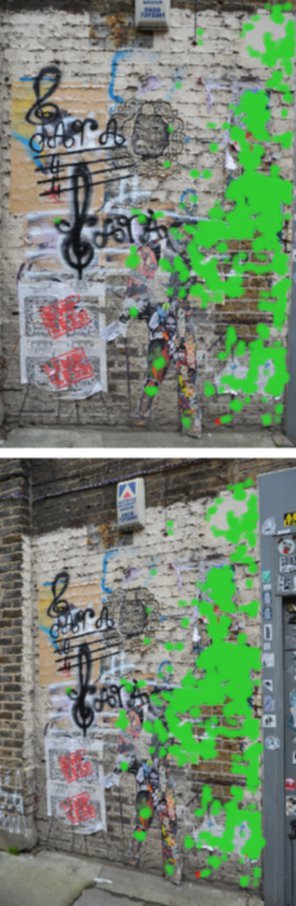} &
    \includegraphics[width=2.3cm]{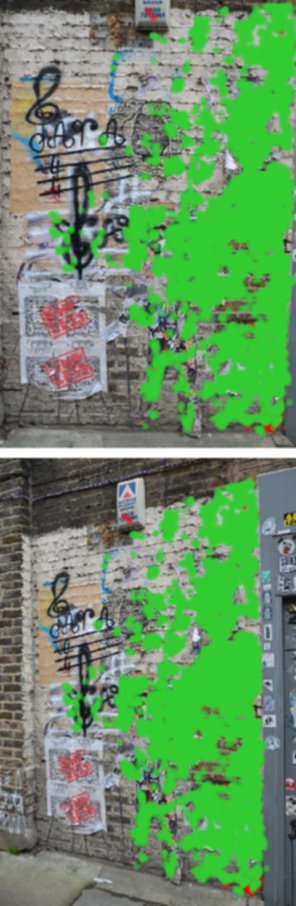} \\[-0.1cm]
    {\smaller 41.6\% (558/1342)} & {\smaller 30.6\% (424/1386)} & {\smaller 72.7\% (722/993)} & {\smaller 99.7\% (1994/2000)} & {\smaller 99.2\% (5952/6000)} \\[0.1cm]
    \smaller SuperPoint~\cite{Detone2018CVPRW} & \smaller D2-Net~\cite{Dusmanu2019CVPR} & \smaller R2D2~\cite{r2d2} & \smaller Sparse-NCNet, 2k & \smaller Sparse-NCNet, 6k\\
    \includegraphics[width=2.3cm]{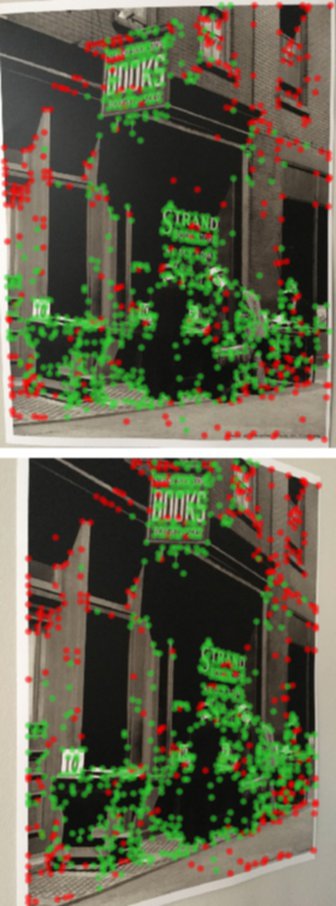} &
    \includegraphics[width=2.3cm]{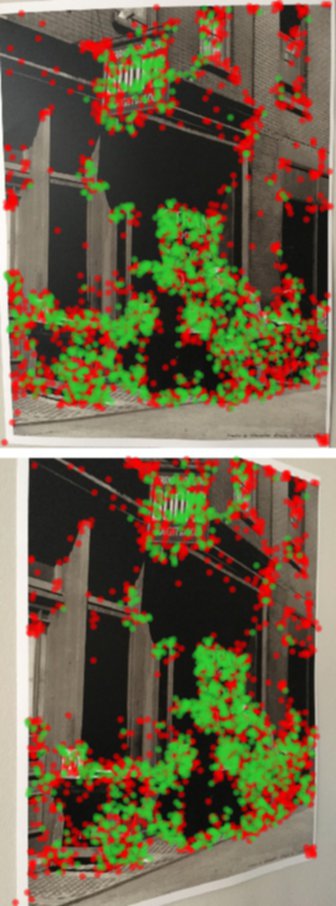} &
    \includegraphics[width=2.3cm]{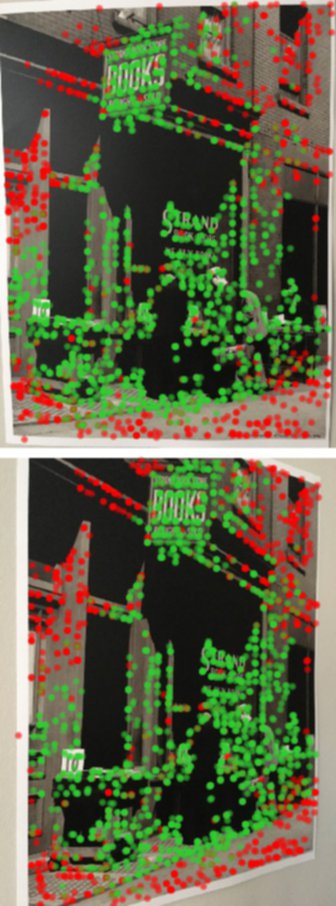} &
    \includegraphics[width=2.3cm]{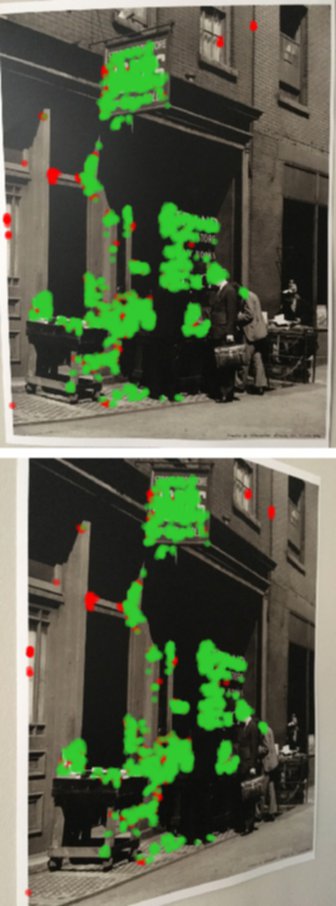} &
    \includegraphics[width=2.3cm]{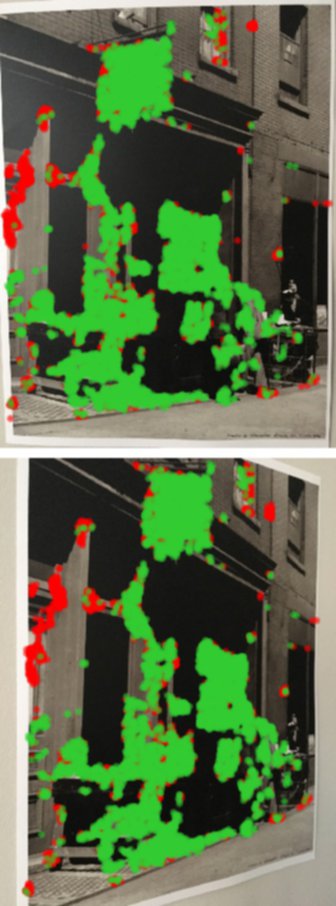} \\[-0.1cm]
    {\smaller 68.8\% (727/1057)} & {\smaller 45.7\% (1170/2561)} & {\smaller 64.8\% (1567/2420)} & {\smaller 85.6\% (1712/2000)} & {\smaller 77.6\% (4656/6000)} \\
    
    \end{tabular}
    \caption{\textbf{HPatches qualitative results (viewpoint).} We present the results of Sparse-NCNet, along with several state-of-the-art methods. The correct correspondences are shown in green, and the incorrect ones in red for a threshold $t=3$px. Below each pair we indicate the fraction of correct matches (both in percentage and absolute values). Our method is presented for both the top 2K matches and the top 6K matches, and it obtains the largest fraction of correct matches for both cases. Examples are from the \emph{viewpoint} sequences.}
    \label{fig:hseq_qual_view}
\end{figure}

\paragraph{Additional qualitative results} are presented in Figures~\ref{fig:hseq_qual_view} and~\ref{fig:hseq_qual_illum}. We compare the MMA of Sparse-NCNet with the state-of-the-art methods SuperPoint~\cite{Detone2018CVPRW}, D2-Net~\cite{Dusmanu2019CVPR} and R2D2~\cite{r2d2}, which are trainable methods for joint detection and description on local features. The correctly matched points are shown in green, while the incorrectly matched ones are shown in red, for a threshold value $t=3$ pixels. For the proposed Sparse-NCNet, results are presented for two different numbers of matches, 2000 and 6000. Results show that our method produces the largest fraction of correct matches, even when considering as many as 6000 correspondences. In particular, note that our method is able to produce a large amount of correct correspondences even under strong illumination changes, as shown in Fig.~\ref{fig:hseq_qual_illum}. Furthermore, note that the nature of the correspondences produced by Sparse-NCNet is different from those of local feature methods. While local feature methods can only produce correspondences on the detected points, which are the local extrema of a particular feature detection function, our method produces densely packed sets of correspondences. This results from Sparse-NCNet's propagation of information in local neighbourhoods, as discussed in Appendix.~\ref{sec:intuitions}. 

\begin{landscape}
\thispagestyle{empty}
\begin{figure}
    \centering
    \vspace*{-1.8em}
    \begin{tabular}{ccccc}
    \smaller SuperPoint~\cite{Detone2018CVPRW} & \smaller D2-Net~\cite{Dusmanu2019CVPR} & \smaller R2D2~\cite{r2d2} & \smaller Sparse-NCNet, 2k & \smaller Sparse-NCNet, 6k \\
    \includegraphics[width=3.4cm]{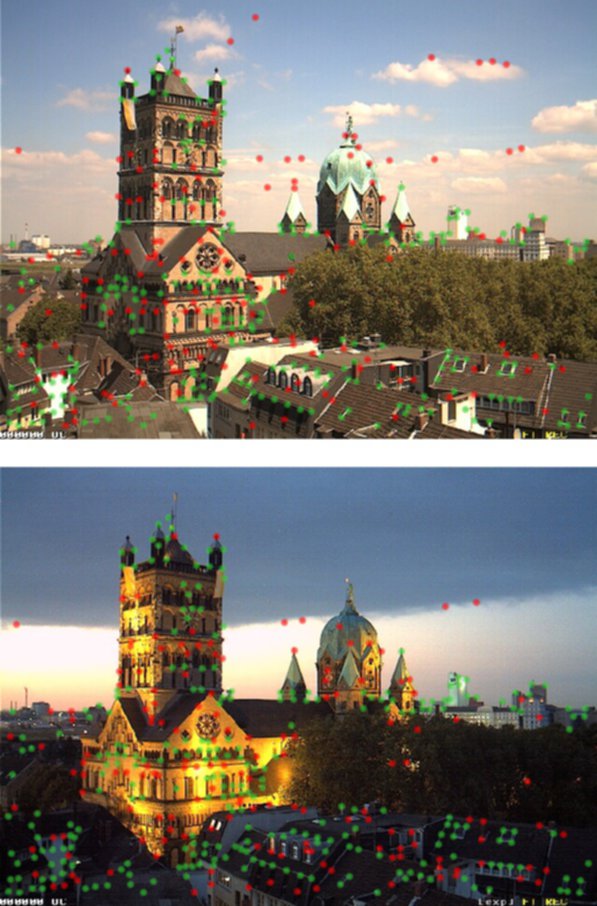} &
    \includegraphics[width=3.4cm]{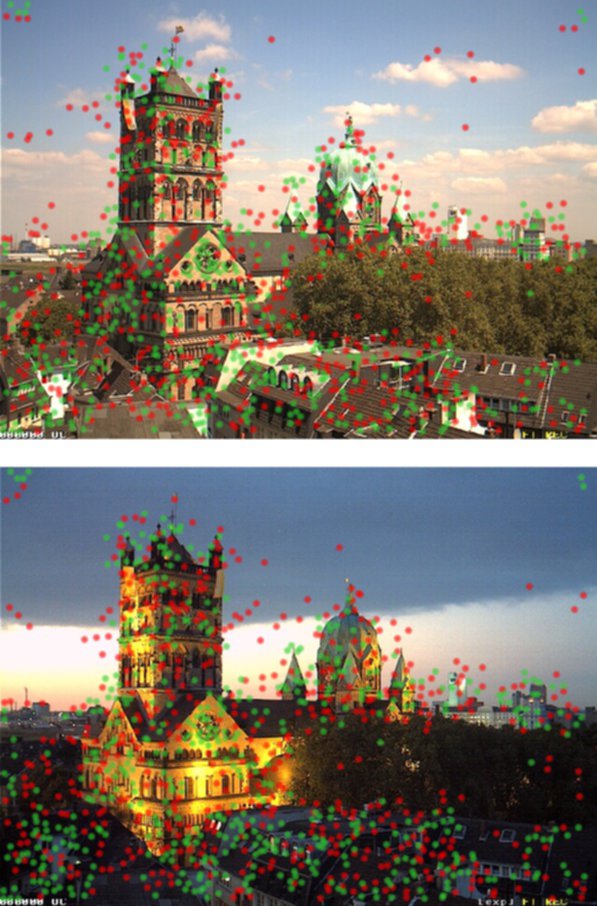} &
    \includegraphics[width=3.4cm]{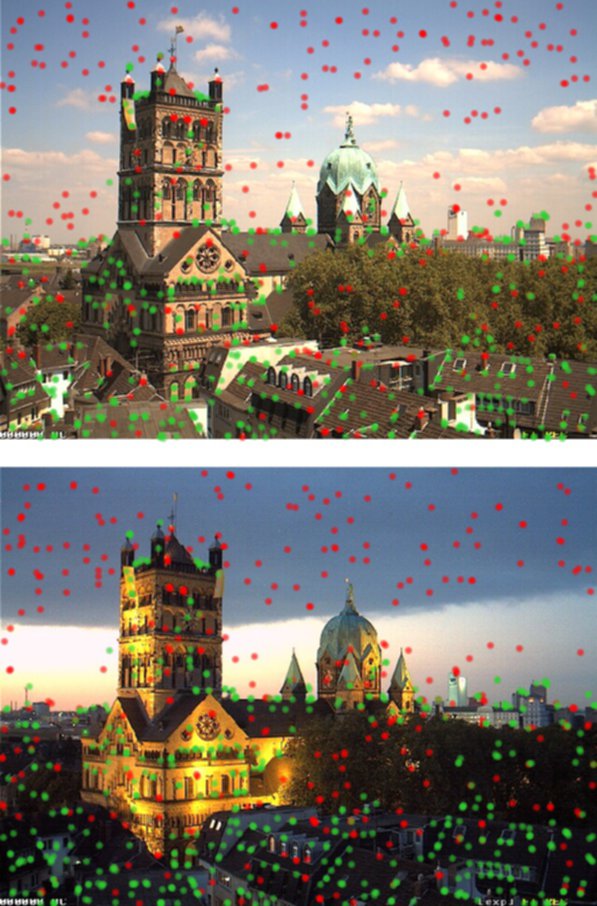} &
    \includegraphics[width=3.4cm]{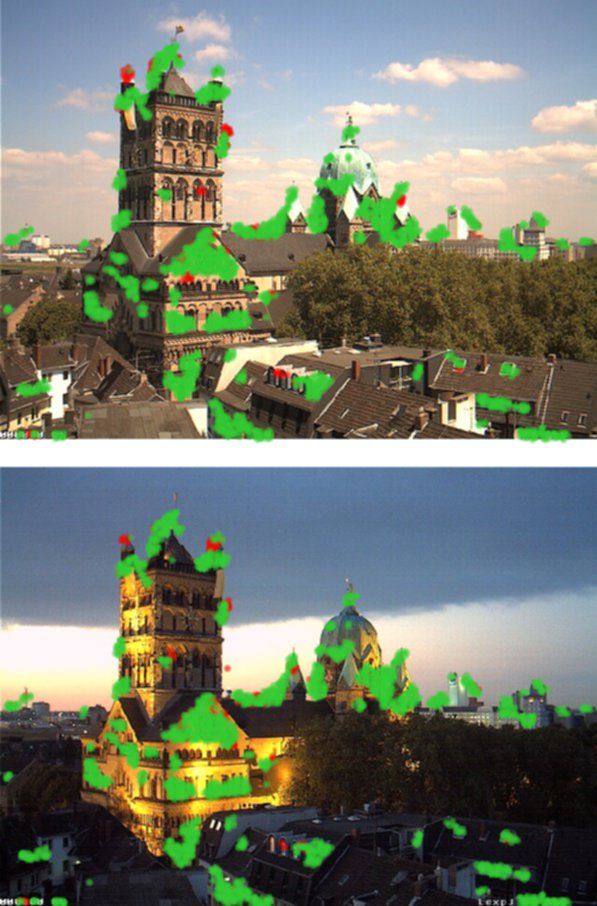} &
    \includegraphics[width=3.4cm]{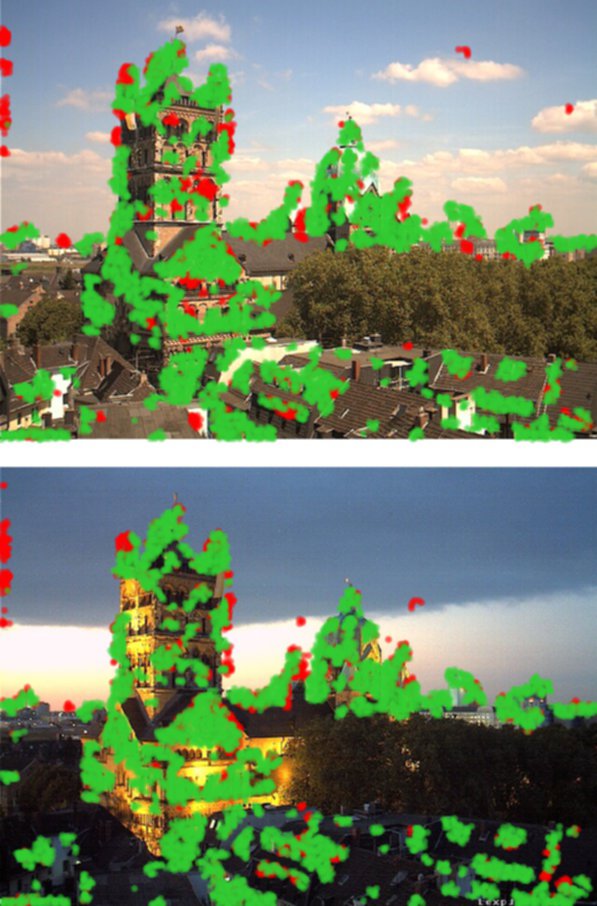} \\[-0.1cm]
    {\smaller 63.0\% (264/419)} & {\smaller 50.8\% (539/1062)} & {\smaller 61.5\% (546/888)} & {\smaller 92.2\% (1844/2000)} & {\smaller 78.9\% (4736/6000)} \\[0.1cm]
    \smaller SuperPoint~\cite{Detone2018CVPRW} & \smaller D2-Net~\cite{Dusmanu2019CVPR} & \smaller R2D2~\cite{r2d2} & \smaller Sparse-NCNet, 2k & \smaller Sparse-NCNet, 6k \\ 
    \includegraphics[width=3.4cm]{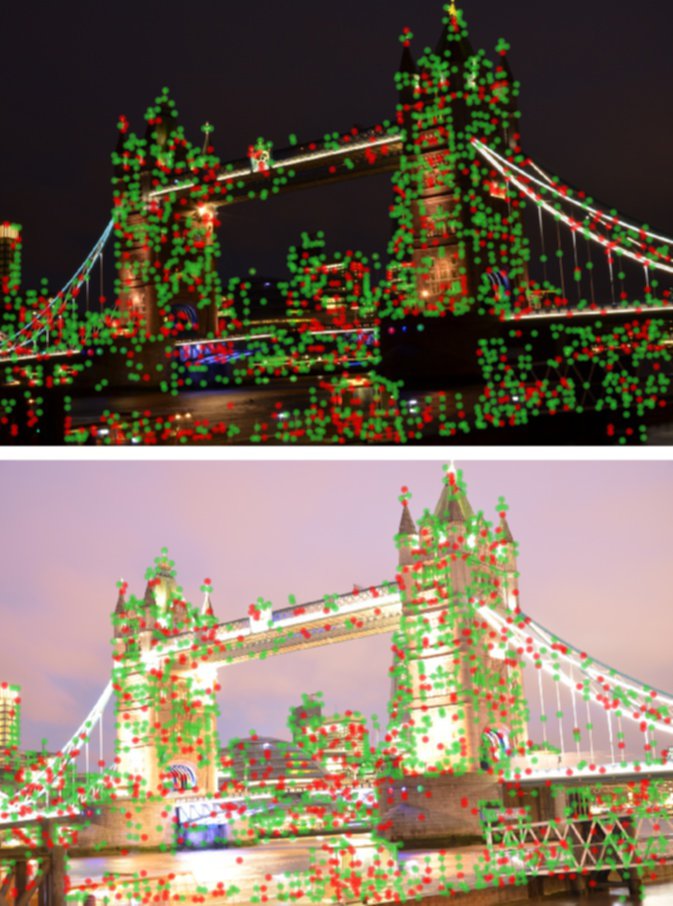} &
    \includegraphics[width=3.4cm]{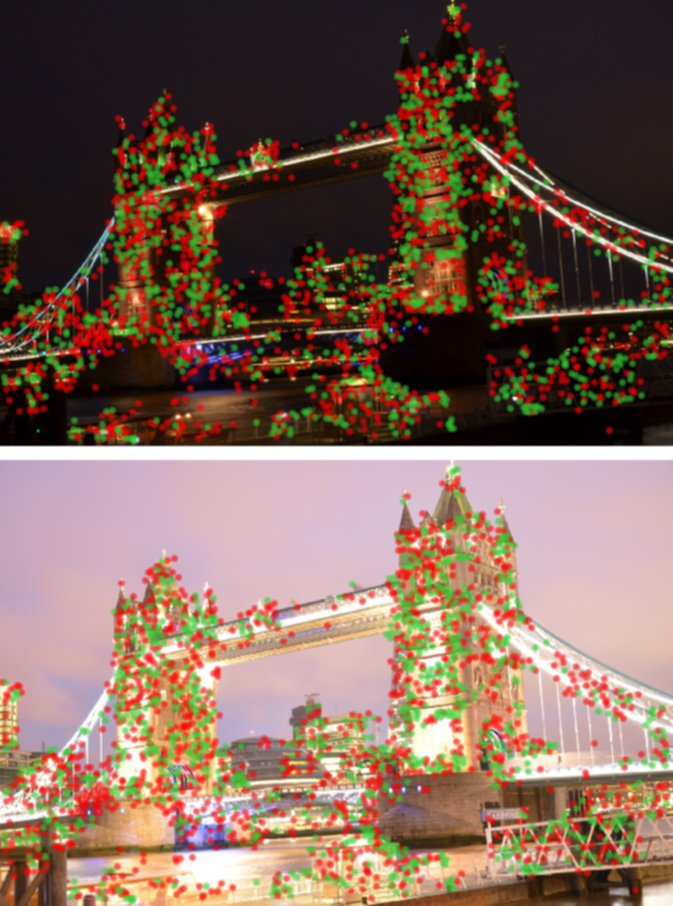} &
    \includegraphics[width=3.4cm]{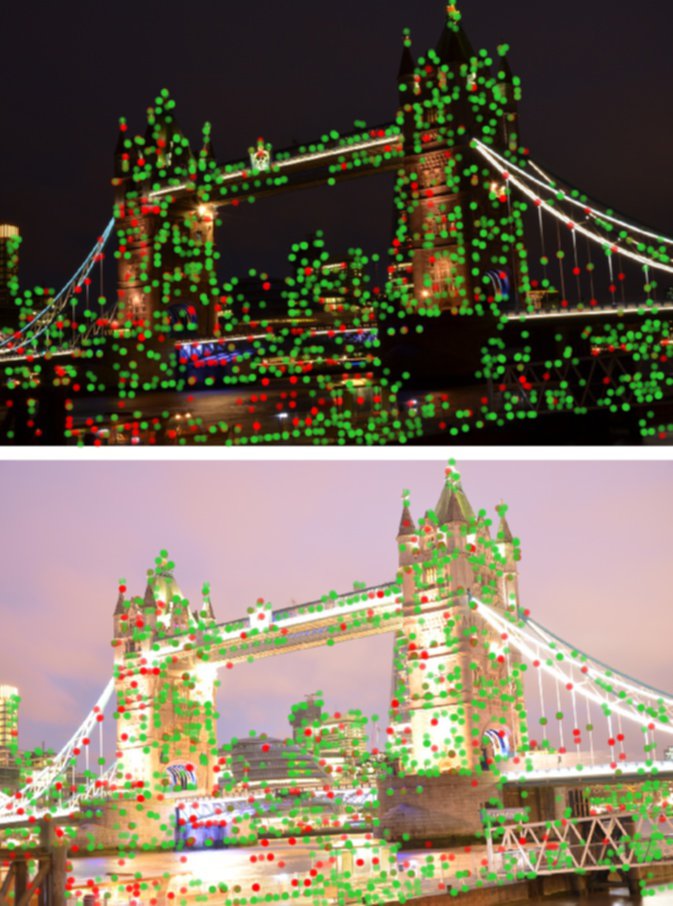} &
    \includegraphics[width=3.4cm]{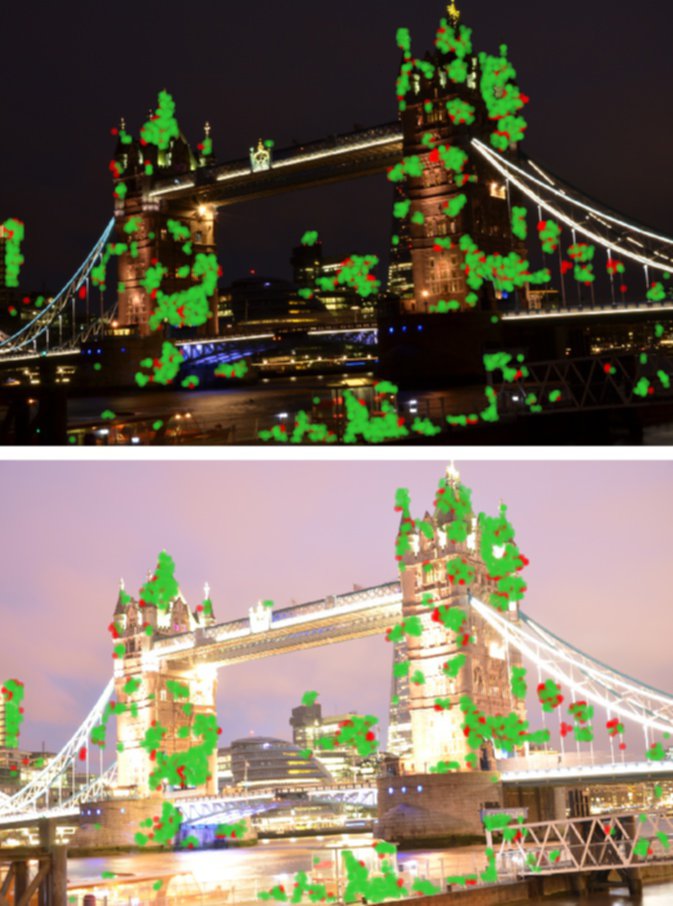} &
    \includegraphics[width=3.4cm]{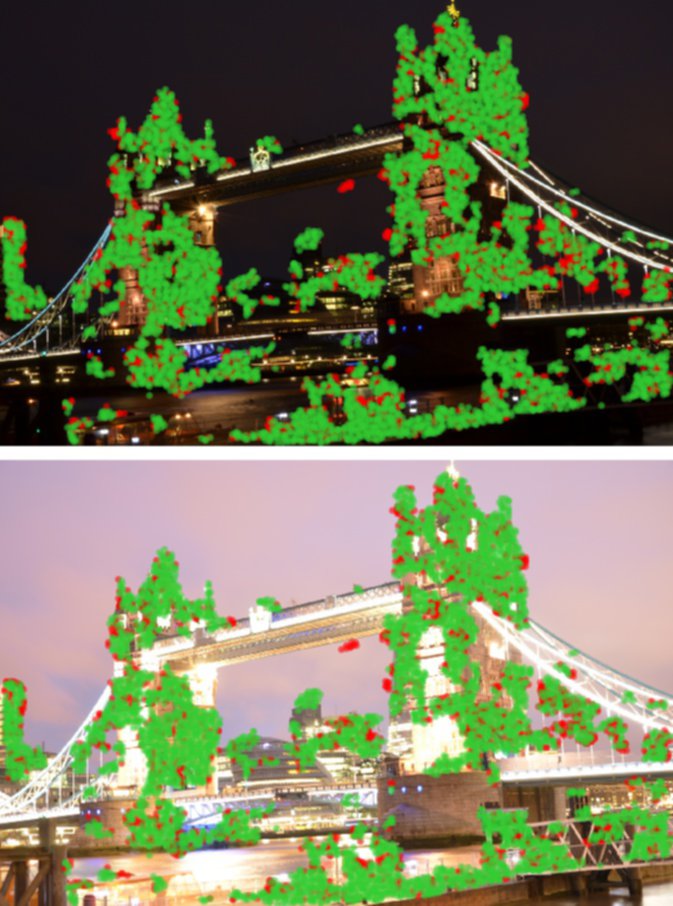} \\[-0.1cm]
    {\smaller 66.6\% (1244/1869)} &  {\smaller 42.6\% (984/2312)} & {\smaller 74.5\% (1667/2238)} & {\smaller 79.9\% (1597/2000)} & {\smaller 78.6\% (4716/6000)}
    \end{tabular}
    \vspace*{-0.5em}
    \caption{\textbf{HPatches qualitative results (illumination).} We present the results of Sparse-NCNet, along with several state-of-the-art methods. The correct correspondences are shown in green, and the incorrect ones in red for a threshold $t=3$px. Below each pair we indicate the fraction of correct matches (both in percentage and absolute values). Our method is presented for both the top 2K and top 6K matches, and it obtains the largest fraction of correct matches for both cases. Examples are from the \emph{illumination} sequences.}
    \label{fig:hseq_qual_illum}
\end{figure}
\end{landscape}

\section{InLoc benchmark\label{sec:inloc_qual}}

We present additional qualitative results from the InLoc indoor localisation benchmark~\cite{Taira18} in Fig.~\ref{fig:inloc}, where the task is to estimate the 6-dof pose of a query image within a large university building. Each image pair is composed of a query image (top row) captured with a cell-phone and a database image (middle row), captured several months earlier with a 3D scanner. Note that the illumination conditions in the two types of images are different. Furthermore, because of the time difference between both images, some objects may have been displaced (\eg furniture) and some aspects of the scene may have changed (\eg wall decoration). For ease of visualisation, we overlay only the top 500 correspondences for each image pair, which appear in green. These correspondences have not been geometrically verified, and therefore contain a certain fraction of incorrect matches. Note however, that most matches are coherent and the few incorrect outliers are likely to be removed when running RANSAC~\cite{Fischler1981} within the PnP pose solver~\cite{gao2003complete}, therefore obtaining a good pose estimate. Also note how Sparse-NCNet is able to obtain correspondences in low textured areas such as walls or ceilings, or on repetitive patterns such as carpets.

\begin{figure}
    \centering
    
    \includegraphics[width=2.8cm]{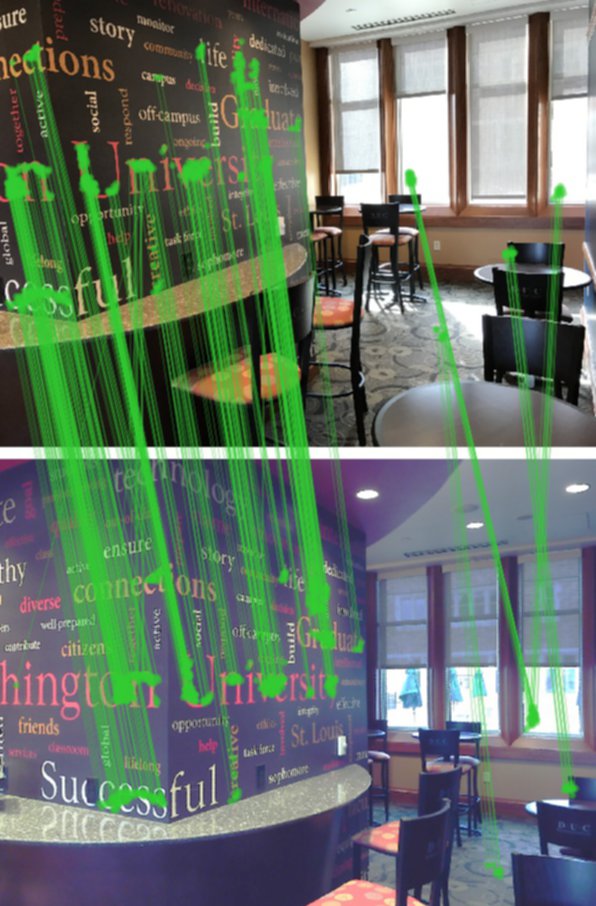}
    \includegraphics[width=2.8cm]{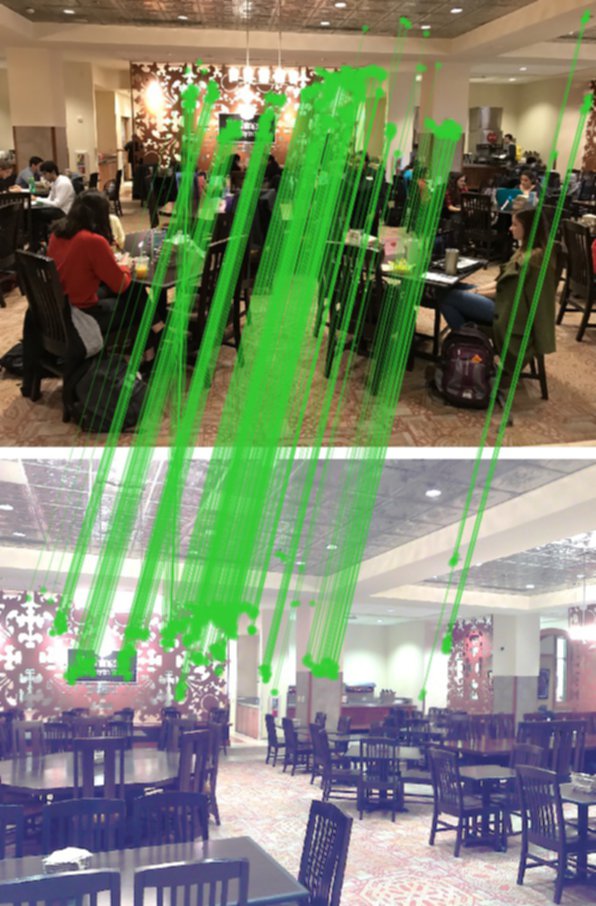} 
    \includegraphics[width=2.8cm]{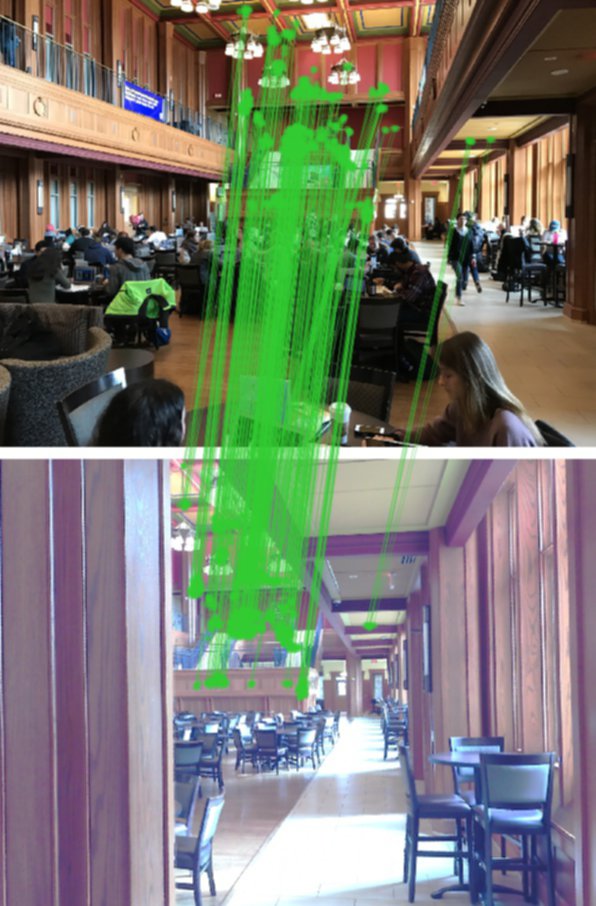} 
    \includegraphics[width=2.8cm]{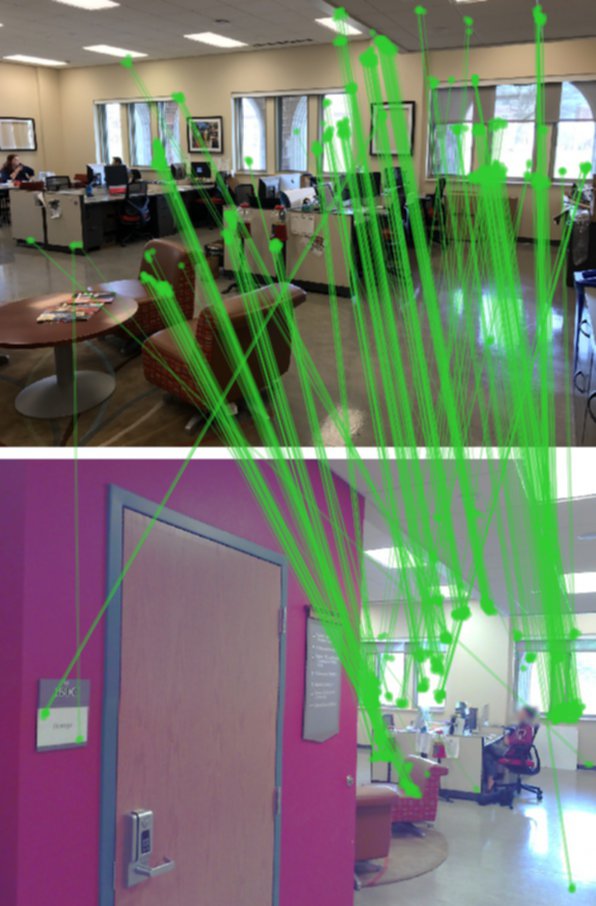} 
    
    \includegraphics[width=2.8cm]{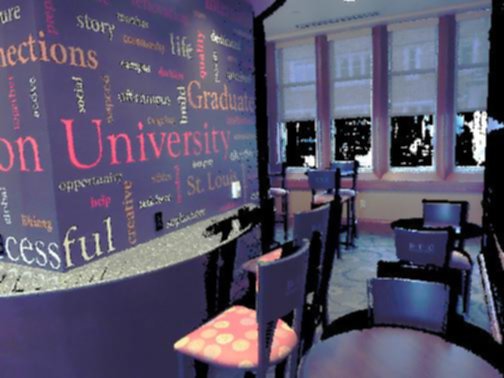}
    \includegraphics[width=2.8cm]{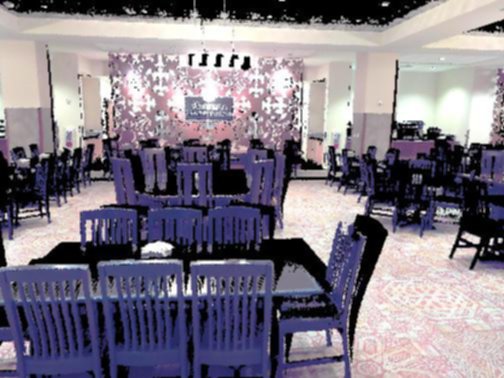} 
    \includegraphics[width=2.8cm]{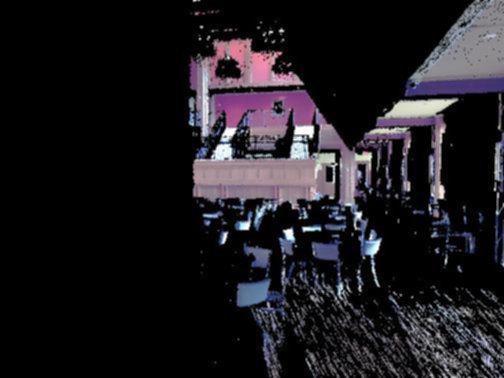} 
    \includegraphics[width=2.8cm]{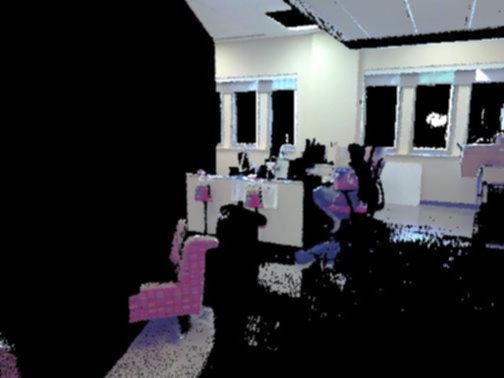} 
    
    \vspace{1em}
    
    \includegraphics[width=2.8cm]{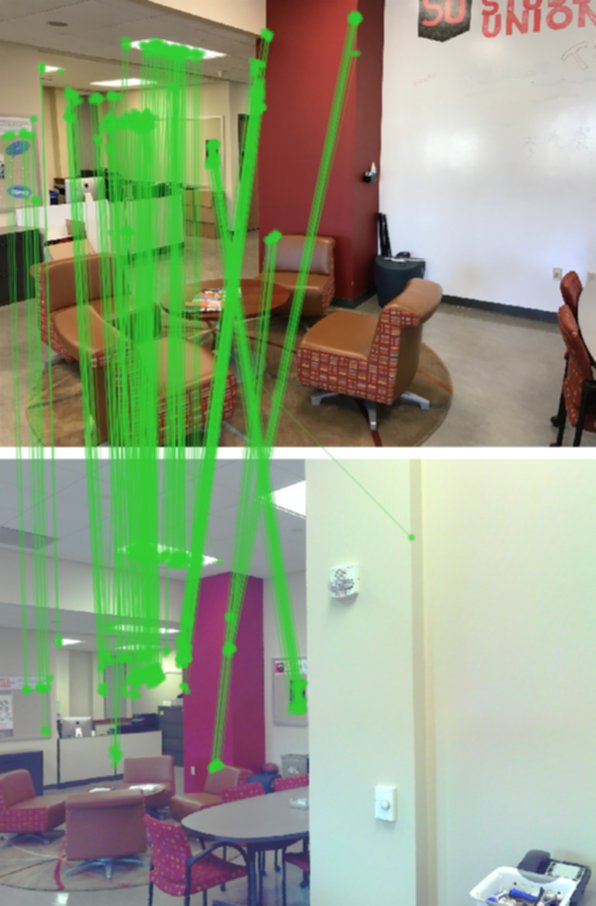}
    \includegraphics[width=2.8cm]{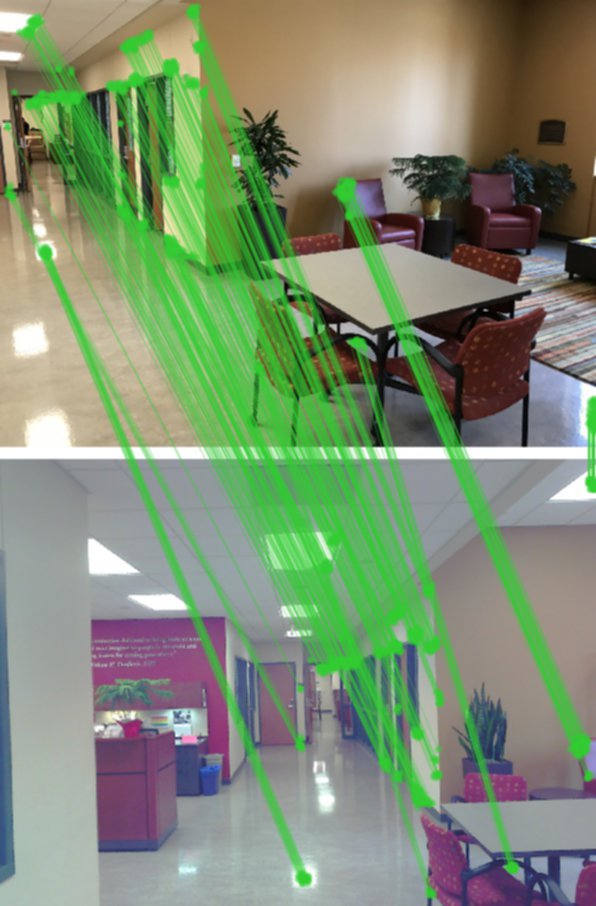} 
    \includegraphics[width=2.8cm]{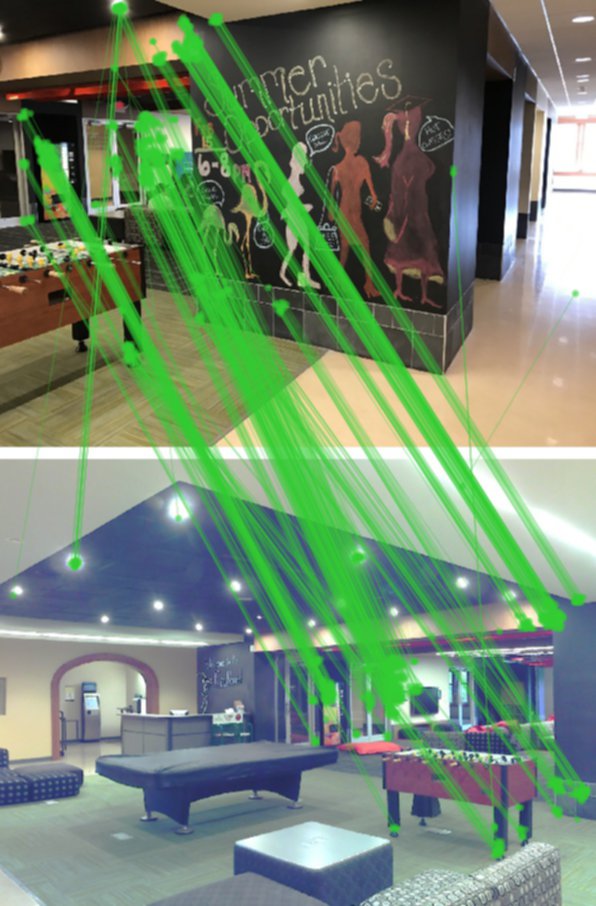} 
    \includegraphics[width=2.8cm]{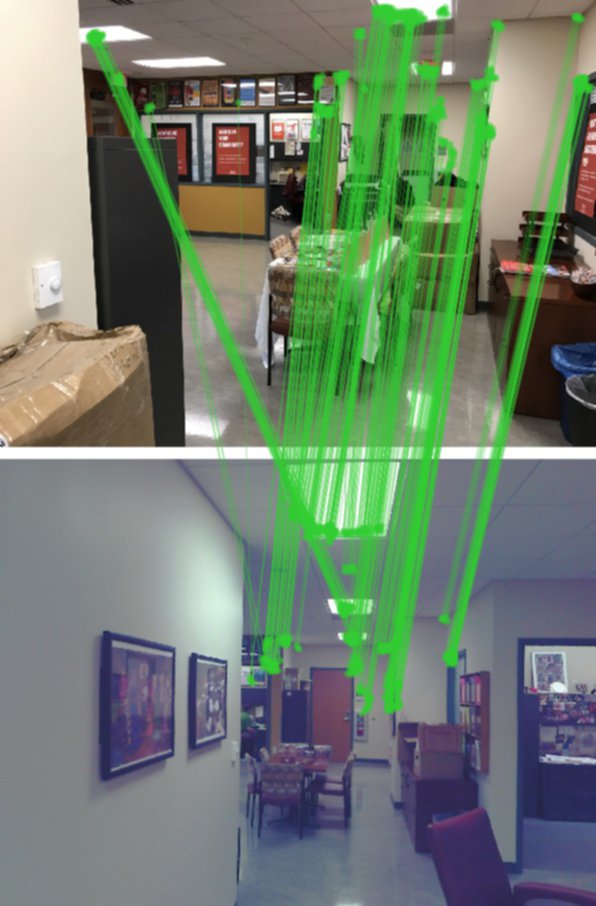} 
    
    \includegraphics[width=2.8cm]{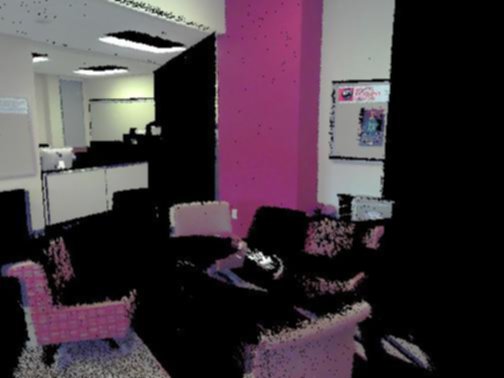}
    \includegraphics[width=2.8cm]{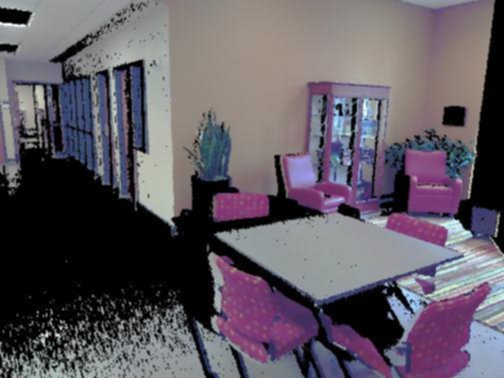} 
    \includegraphics[width=2.8cm]{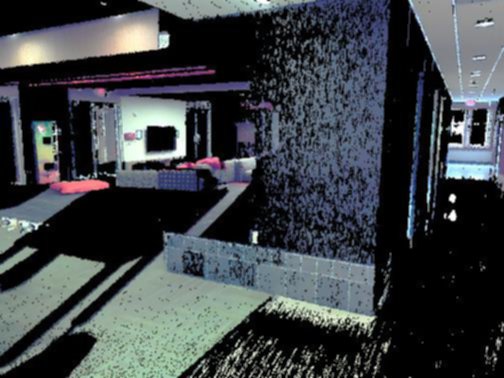} 
    \includegraphics[width=2.8cm]{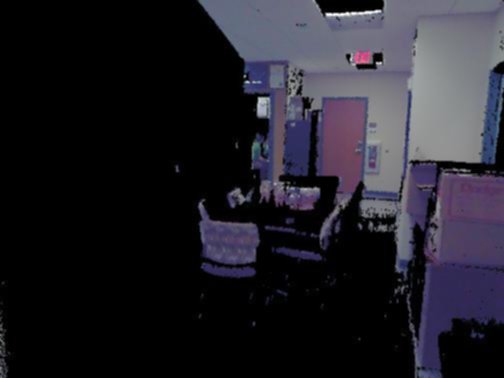} 
    
    \caption{\textbf{InLoc qualitative results.} For each image pair, we show the top 500 matches produced by Sparse-NCNet between the query image (top row) and database image (middle row). In addition we show the rendered scene from the estimated query 6-dof pose (bottom row), obtained by running RANSAC+PnP\cite{Fischler1981,gao2003complete} on our matches. Note these rendered images are well aligned with the query images, demonstrating that the estimated poses have low translation and rotation errors.}  
    \label{fig:inloc}
\end{figure}

\FloatBarrier

\section{Aachen day-night benchmark\label{sec:aachen_qual}}

\begin{figure}
    \centering
    \includegraphics[width=3.6cm]{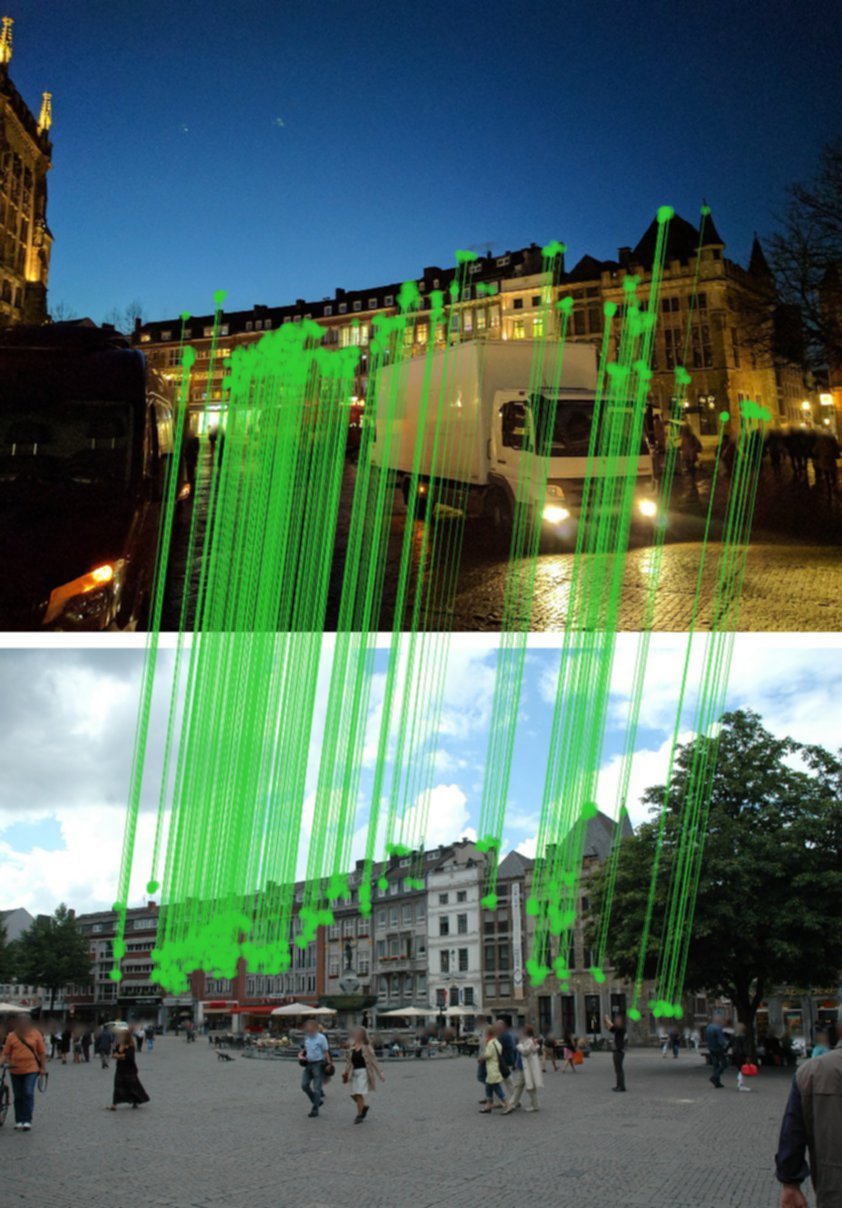}
    \includegraphics[width=3.6cm]{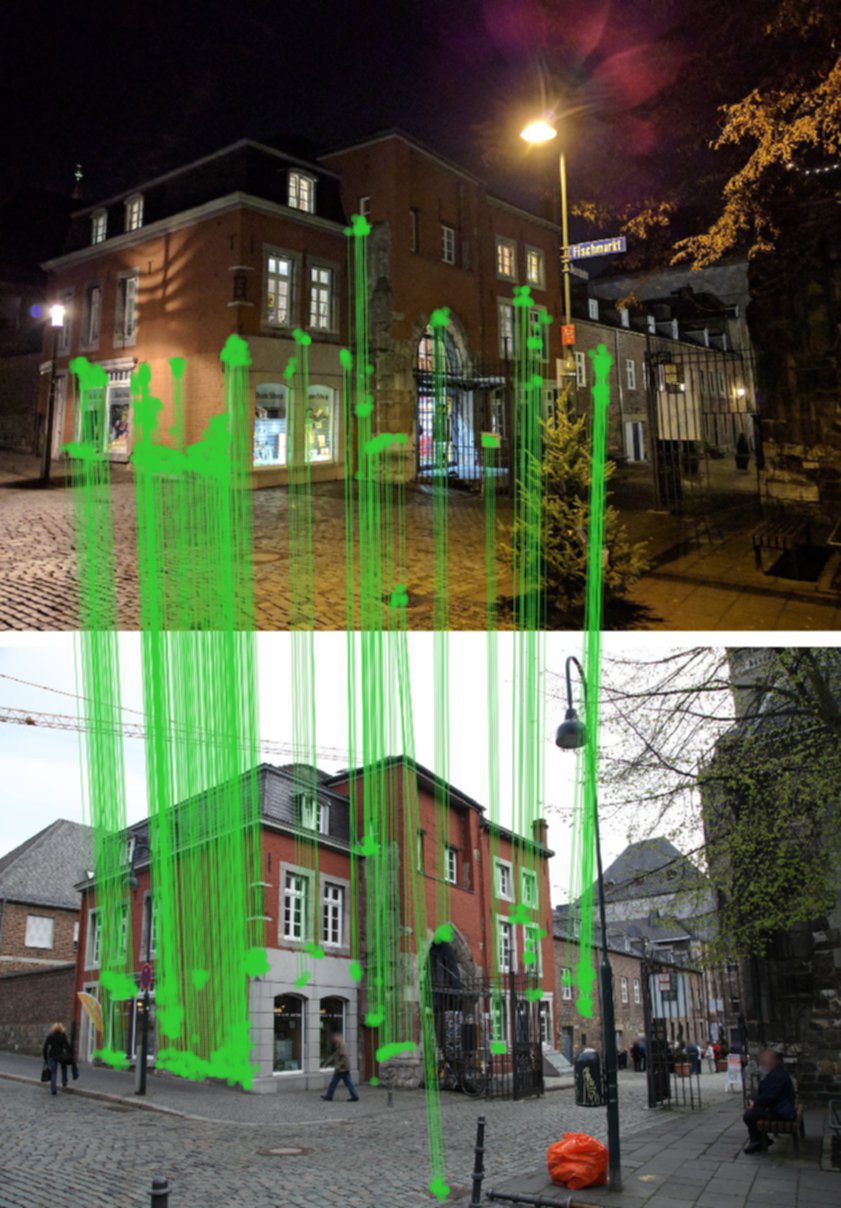}
    \includegraphics[width=3.6cm]{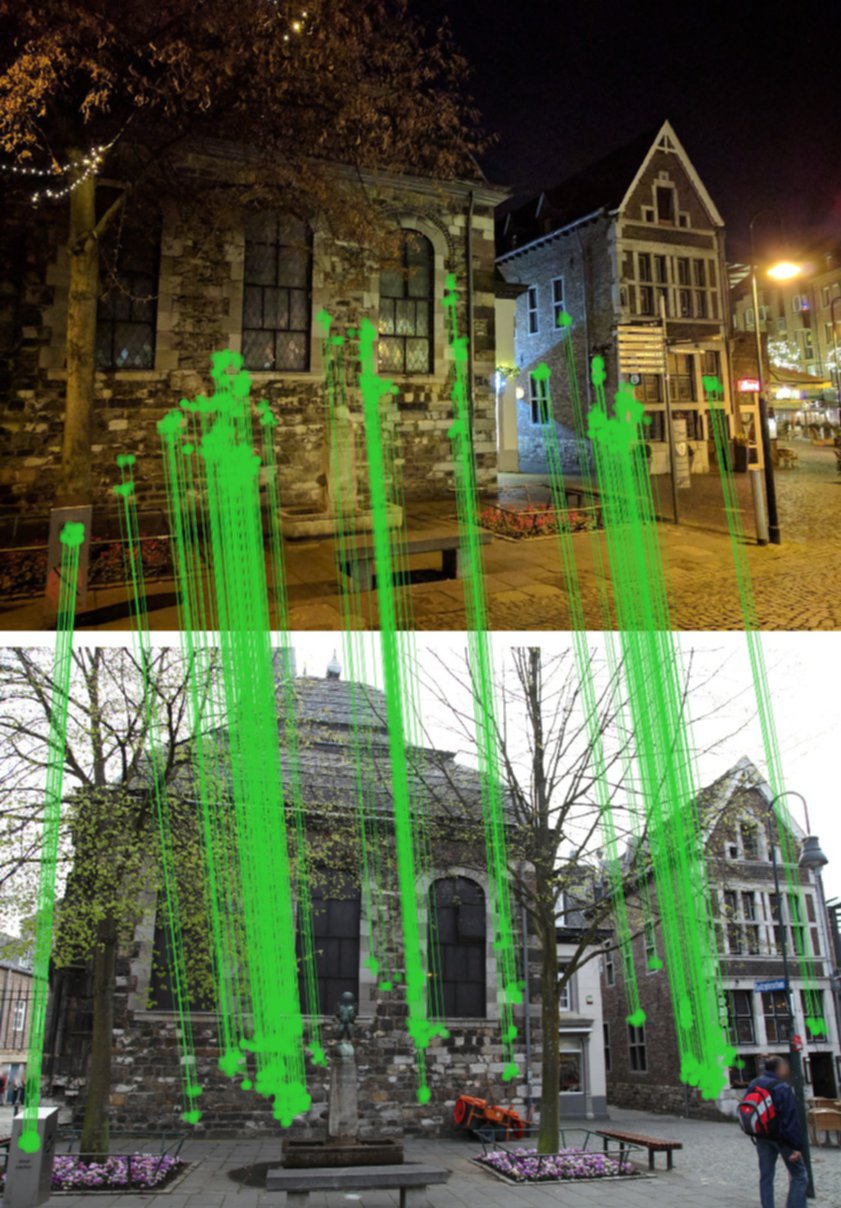}
    
    \includegraphics[width=3.6cm]{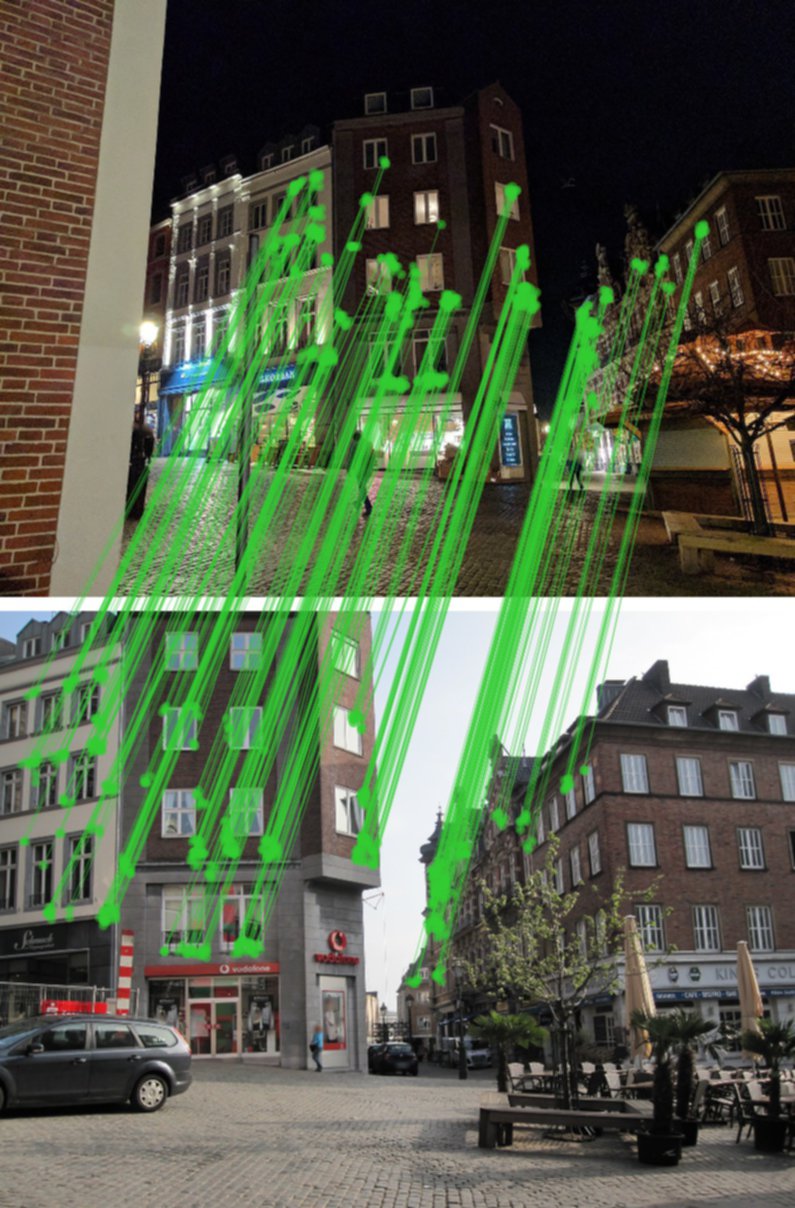}
    \includegraphics[width=3.6cm]{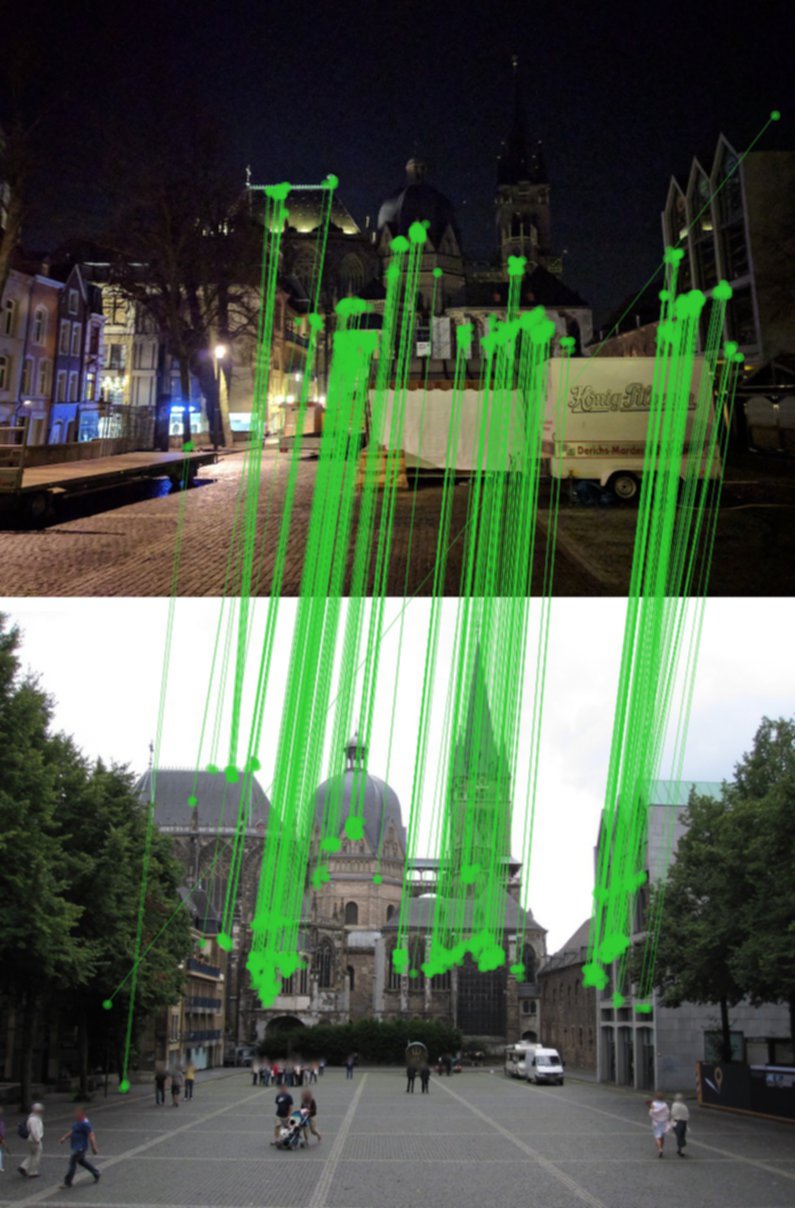}
    \includegraphics[width=3.6cm]{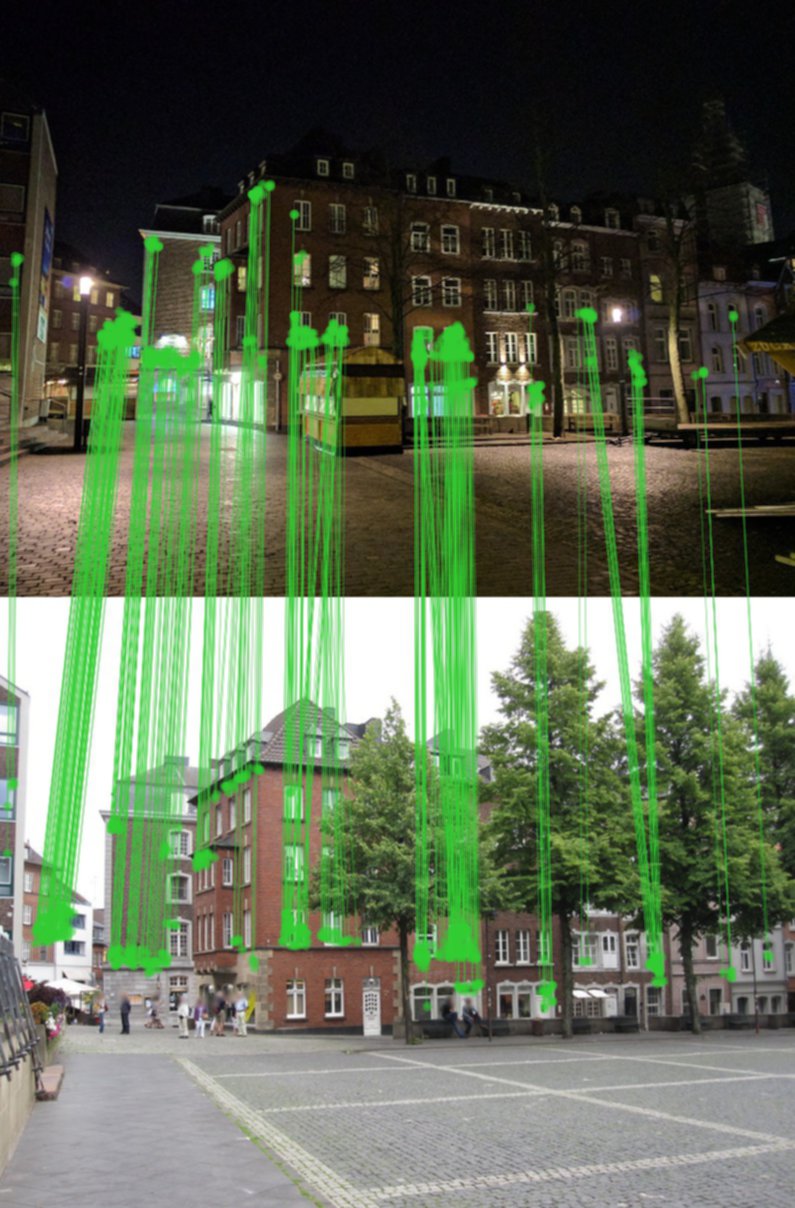}
    
    \includegraphics[width=3.6cm]{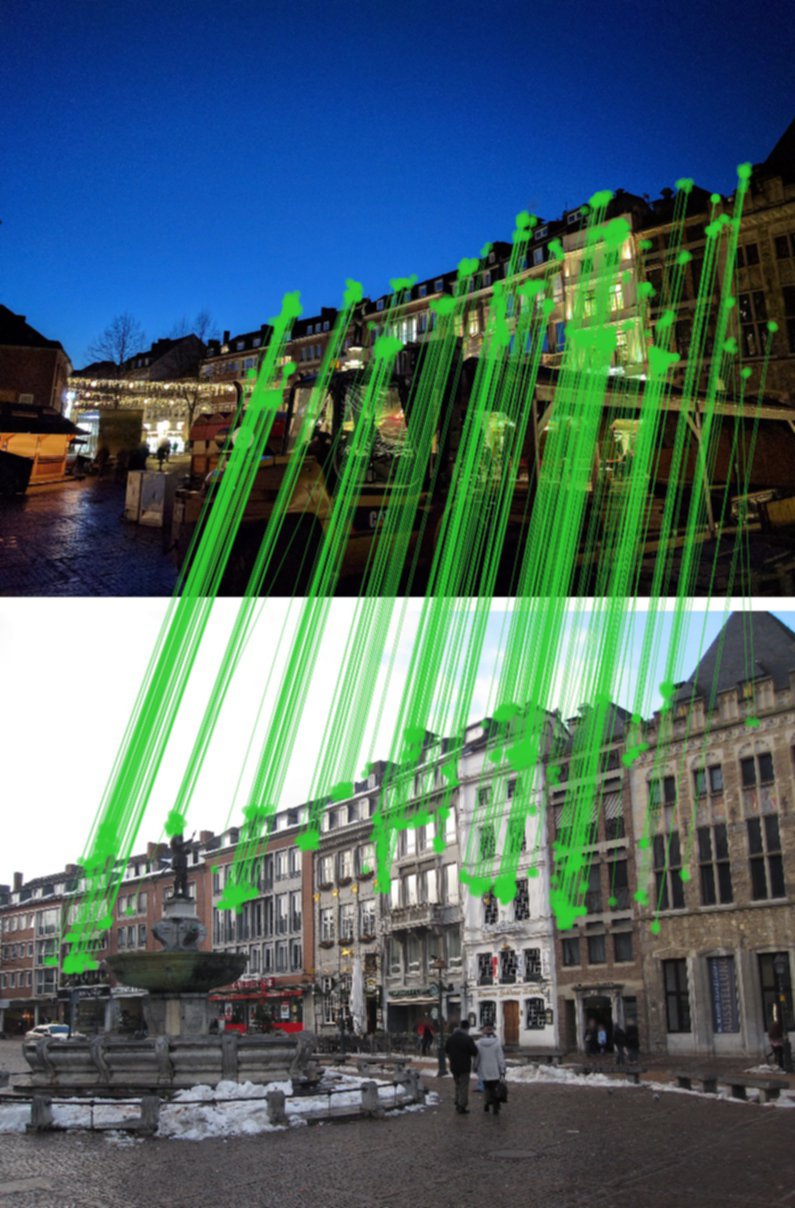}
    \includegraphics[width=3.6cm]{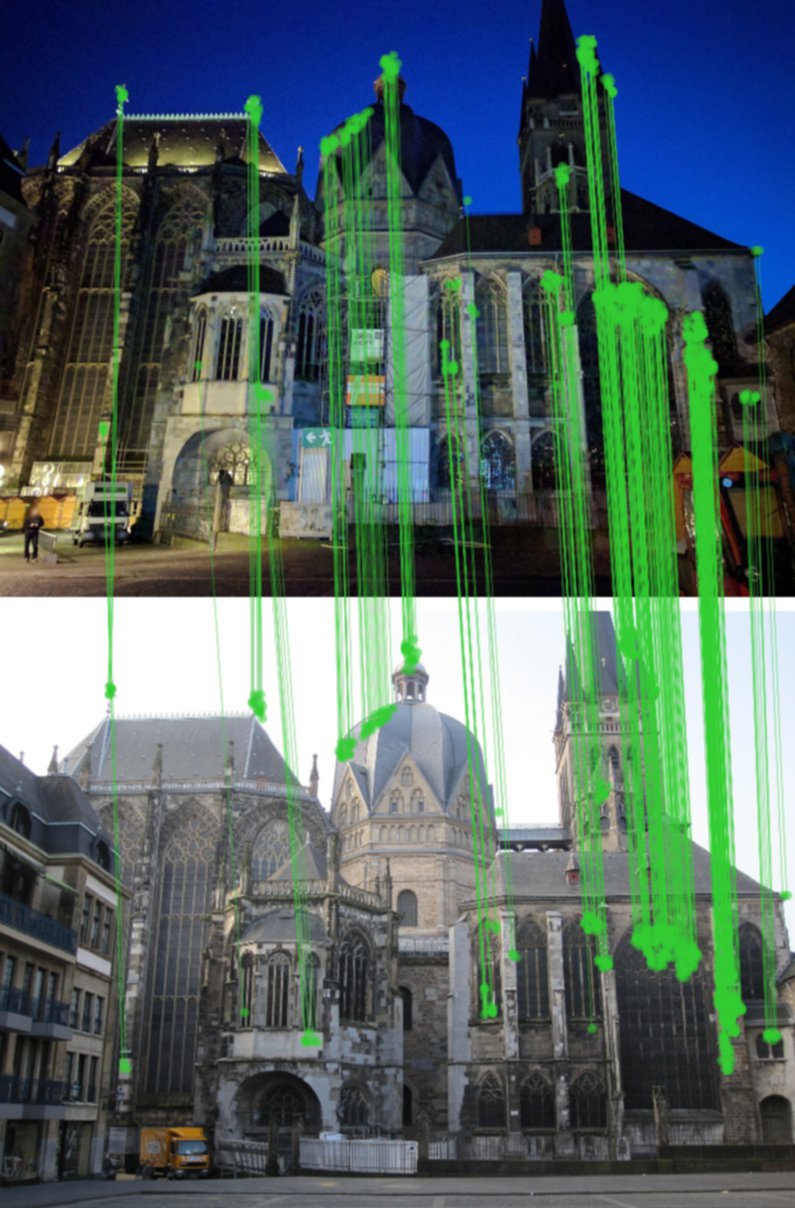}
    \includegraphics[width=3.6cm]{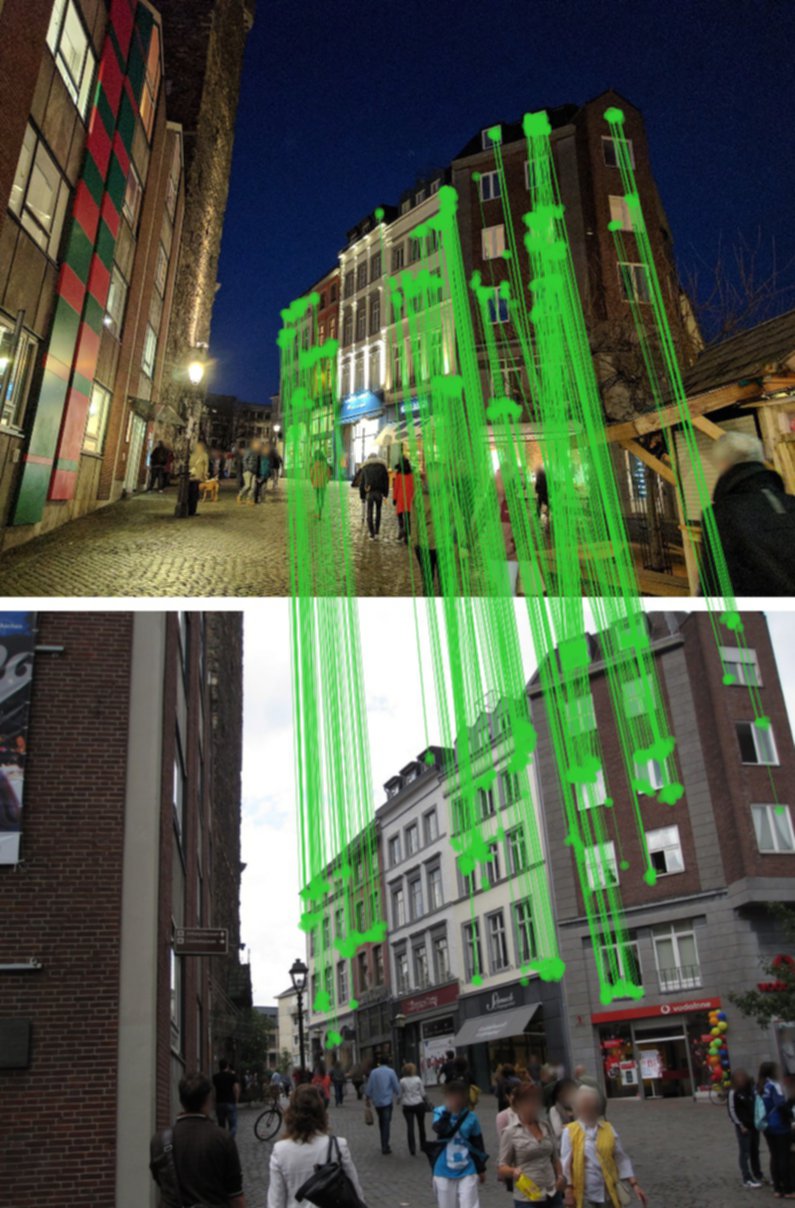}
    \caption{\textbf{Aachen day-night results.}  We show the top 500 correspondences obtained by Sparse-NCNet between the night query image (top) and the database day image (bottom). Note that the large majority of matches are correct, despite the strong illumination changes.}
    \label{fig:aachen_qual}
\end{figure}

Additional qualitative results from the Aachen-day benchmark are shown in Fig.~\ref{fig:aachen_qual}. We show several image pairs composed of night query images (top) and their top matching database images (bottom), according to the average matching score of Sparse-NCNet. For each image pair, we overlay the top 500 correspondences obtained with Sparse-NCNet. Note that these correspondences were not geometrically verified by any means. Nevertheless, as seen in Fig.~\ref{fig:aachen_qual}, most correspondences are coherent and seem to be correct, despite the strong changes in illumination between night and day images.

\FloatBarrier

\end{document}